\documentclass[11pt]{article}

\usepackage[preprint]{neurips_2026}

\usepackage[utf8]{inputenc} 
\usepackage[T1]{fontenc}    
\usepackage{hyperref}       
\usepackage{url}            
\usepackage{booktabs}       
\usepackage{nicefrac}       
\usepackage{microtype}      
\usepackage{xcolor}         

\usepackage{amsmath,amssymb,amsthm,amsfonts}
\usepackage{mathtools}
\usepackage{bm}
\usepackage{graphicx}
\usepackage{multirow}
\usepackage{algorithm}
\usepackage{algpseudocode}

\newcommand{\E}{\mathbb{E}}

\newcommand{\KL}{\mathrm{KL}}
\newcommand{\N}{\mathcal{N}}
\newcommand{\R}{\mathbb{R}}
\newcommand{\bx}{\bm{x}}
\newcommand{\bw}{\bm{w}}
\newcommand{\bmu}{\bm{\mu}}
\newcommand{\bSigma}{\bm{\Sigma}}
\newcommand{\bLambda}{\bm{\Lambda}}
\newcommand{\btheta}{\bm{\theta}}

\newcommand{\bQ}{\bm{Q}}

\newcommand{\bXi}{\bm{\Xi}}
\newcommand{\bI}{\bm{I}}

\title{Variational Filtering for Bayesian Last Layers with an Independent Binary Probit Surrogate}

\author{%
  Dimitrije Markovi\'c \thanks{homepage: \url{https://di.markov.icu}} \\
  Independent Researcher\\
  Vienna, Austria\\
  \texttt{research@markov.icu} \\
  \And
  Conor Heins \\
  Independent Researcher\\
  Berlin, Germany \\
  \texttt{conor.heins@gmail.com} \\
  \And
  Alexander Tschantz \\
  University of Sussex \\
  Brighton, UK \\
  \texttt{tschantz.alec@gmail.com} \\
  \And
  Christopher L Buckley  \\
  Department of Informatics \\ University of Sussex \\
  Brighton, UK \\
  \texttt{c.l.buckley@sussex.ac.uk} \\
}

\begin{document}

\maketitle

\begin{abstract}
Modern deep classifiers trained by gradient descent need many passes over the data to become accurate, and even then are not always well calibrated; post-hoc temperature scaling corrects calibration after training but cannot influence training-time accuracy or convergence. We propose a variational filtering algorithm for the last layer of deep classifiers, whose parameters are treated as the latent state of a linear-Gaussian state-space model. Each mini-batch is assimilated by an exact prediction step followed by a closed-form variational update, made conjugate by an independent binary probit (IBProbit) surrogate for the categorical likelihood. The filter's process noise stabilises the updates under feature drift and keeps the predictive distribution from collapsing to a point estimate, allowing the same procedure to be used for linear probing on a frozen backbone and for full-network fine-tuning. We evaluate on DINOv3 backbones against two baselines tuned under an identical protocol: AdamW with cross-entropy and post-hoc temperature scaling, and Variational Bayesian Last Layers (VBLL). As a linear probe on frozen features, IBProbit attains the best accuracy and negative log-likelihood on all five downstream tasks at the largest backbone, and on two of the five at the smallest. In out-of-distribution detection with CIFAR probes on the largest backbone, IBProbit reduces the false-positive rate at 95\% true-positive rate by a factor of 1.4 to 3.2 relative to the better of the two baselines, on all three dataset pairs. When the whole network is fine-tuned on ImageNet-1k, IBProbit reaches the final performance of the cross-entropy baseline in half the wall-clock time.
\end{abstract}

\section{Introduction}

Modern deep neural networks, despite achieving remarkable predictive accuracy, are often weakly calibrated, as their confidence scores do not reliably reflect the true probability of correctness \citep{guo2017calibration}. This matters in safety-critical applications such as medical diagnosis, autonomous driving, and financial decision-making, where downstream decisions depend on predicted probabilities, and not just the correctness of the predicted labels. Temperature scaling \citep{guo2017calibration} is the standard post-hoc correction and is a strong in-distribution baseline, but it is a one-parameter correction applied after training on a separate validation split; it cannot influence the learned features, and it offers no mechanism to keep a model calibrated \emph{during} training or fine-tuning, where the feature distribution is itself shifting.

Bayesian approaches to deep learning offer a principled alternative, replacing a point estimate of the weights with a distribution and propagating the associated uncertainty into predictions \citep{blundell2015weight, izmailov2021bayesian}. Full Bayesian inference over all network parameters remains computationally prohibitive at scale. The ``Bayesian last-layer'' (BLL) paradigm is a practical compromise: only the final classification layer is treated probabilistically, while the feature extractor remains deterministic \citep{kristiadi2020being, daxberger2021laplace}. This preserves the representational power of large pretrained backbones while adding meaningful epistemic uncertainty at a small fraction of the compute cost. Existing sampling-free BLL methods either rely on softmax-bound ELBOs that still require gradient-based head optimization (VBLL \citep{harrison2024variational}) or on post-hoc Laplace approximations whose uncertainty quality is bounded by the converged point estimate \citep{daxberger2021laplace}.

We build on \citet{wojnowicz2022}, who introduce \emph{categorical-from-binary} (CB) likelihoods: categorical models whose likelihood is lower-bounded by a product of $K$ independent per-class binary likelihoods evaluated at the one-hot encoding of the label, so that this independent binary (IB) model can serve as a tractable surrogate for posterior inference. The surrogate model admits a conjugate Gaussian update that we can solve in closed form via coordinate-ascent variational inference (CAVI), with no gradient-based optimization of the classification head. We extend this approach in two ways that are relevant in practice. First, we turn the batch update into a \emph{variational filter}: the head is the latent state of a linear-Gaussian state-space model, and each mini-batch is one exact prediction step followed by a closed-form variational update. The process noise of the prediction step serves two purposes at once --- it tracks changes in the optimal head as the backbone is fine-tuned, and it counteracts covariance contraction as evidence accumulates; enabling a well calibrated posterior predictive data likelihood. Other approaches, like streaming variational Bayes \citep{broderick2013streaming}, accumulate evidence without discounting it. Here, instead we introduce a single scalar hyper-parameter that sets a covariance floor for arbitrarily long training. In addition, we exploit the fact that the independent-probit likelihood induces a precision matrix that is \emph{shared across all $K$ classes}, so two Cholesky factorisations per batch (not per CAVI step) serve all $K$ class-conditional updates, and the $T$-iteration CAVI inner loop itself reduces to matrix multiplications with no further matrix solves or inversions.

Concretely, we ask: can a closed-form variational filter over the last-layer weights (i) match an extensively trained cross-entropy and temperature-scaling baseline on both accuracy and calibration in fewer epochs, (ii) surpass the leading sampling-free Bayesian last-layer alternative on the same backbone, and (iii) scale to ImageNet-scale probing on modern self-supervised backbones? Our contributions are: (i) a variational filtering algorithm for the last layer, pairing a Kalman-style prediction step with a closed-form CAVI update under the independent binary probit (IBProbit) surrogate; (ii) a parallel-in-$K$, Cholesky-preconditioned inner loop accelerated by parameter expansion (Algorithm~\ref{alg:cavi}); (iii) a comparison under a matched training protocol against AdamW+cross-entropy with temperature scaling and against Variational Bayesian Last Layers \citep{harrison2024variational}, where IBProbit leads on accuracy and NLL at every task at the largest backbone, and on more tasks as the backbone size increases; (iv) out-of-distribution detection with CIFAR-10 and CIFAR-100 probes on the largest backbone, where IBProbit has the lowest false-positive rate at 95\% true-positive rate on all three dataset pairs (lower than the better baseline by a factor of about three on CIFAR-10 vs SVHN) and loses no in-distribution accuracy; and (v) full-network fine-tuning of DINOv3-Huge on ImageNet-1k, where IBProbit reaches the final accuracy, NLL and Brier score of the cross-entropy baseline in half the wall-clock time, and VBLL does not reach them within the same budget.

Code and scripts to reproduce all results are released at the following GitHub repository \url{https://github.com/dimarkov/bllarse}. 

\section{Related Work}

\paragraph{Calibration of deep classifiers.}
Post-hoc temperature scaling \citep{guo2017calibration} is a strong in-distribution baseline but, being applied after training, cannot influence training-time convergence or accuracy. Training-time alternatives reshape the loss --- focal loss \citep{mukhoti2020calibrating}, mixup \citep{thulasidasan2019mixup} --- and recent architecture-driven gains in calibration are uneven and architecture-specific \citep{minderer2021revisiting}. We instead replace the softmax likelihood with a categorical-from-binary probit likelihood \citep{wojnowicz2022} and infer the last-layer weights through its independent binary surrogate.

\paragraph{Bayesian Last Layers.}
Treating only the final layer as Bayesian is a long-established compromise: a single Bayesian layer fixes overconfidence far from training data \citep{kristiadi2020being}, and an additional infinite-feature GP extension is needed for cubic-rate asymptotic calibration \citep{kristiadi2021infinite}. For classification, Variational Bayesian Last Layers (VBLL) \citep{harrison2024variational} keep the categorical likelihood tractable via a double-Jensen log-sum-exp lower bound on the ELBO; for regression, \citet{fiedler2023improved} proposes a related Bayesian last-layer method via a reformulated marginal likelihood. We differ in two ways. First, we bound a different quantity: instead of keeping the softmax and bounding its ELBO, we change the \emph{likelihood} to a categorical-from-binary probit and lower-bound it by its independent binary probit surrogate \citep{wojnowicz2022}, whose ELBO is maximized by a \emph{closed-form} conjugate update rather than by gradient steps. Second, we apply this update as a recursive Bayesian filter over a drifting head.

\paragraph{Filtering the last layer.}
Streaming variational Bayes \citep{broderick2013streaming} updates a variational posterior recursively over the data, with a static parameter, so its precision accumulates without bound; adding dynamics to the parameter yields a filter. Closest to our work, \citet{titsias2024kalman} place a state-space model over the last-layer weights and track it with a Kalman filter whose covariance is, as here, shared across classes. They handle classification by replacing the softmax with a Gaussian likelihood on the one-hot labels, and predict by Monte Carlo through the exact softmax with a gradient-fitted calibration parameter. Our update keeps a non-Gaussian probit observation model and solves it variationally: it has the same form as theirs, with the Albert--Chib expectations in place of the one-hot targets, and its predictive is closed-form. We also target a different regime, calibration under standard probing and fine-tuning rather than online continual learning.

\paragraph{Other uncertainty-quantification approaches.}
Whole-network UQ methods --- deep ensembles \citep{lakshminarayanan2017simple}, MC Dropout \citep{gal2016dropout}, SWAG \citep{maddox2019simple} --- are more expressive but substantially more expensive at train and inference time, and their absolute calibration still degrades under shift, with deep ensembles among the most robust but still imperfect \citep{snoek2019can}; post-hoc Laplace approximations \citep{daxberger2021laplace} avoid that cost but, being a Gaussian centered at the trained MAP, inherit any pathologies of the converged point estimate. \citet{harrison2024variational} found variational last-layer methods competitive with these baselines at a fraction of the compute, so we focus our direct comparison on the VBLL family.

\paragraph{Pólya-Gamma augmentation.}
Pólya-Gamma augmentation \citep{polson2013bayesian} yields conjugate Gaussian updates for logistic regression, and stick-breaking variants \citep{linderman2015dependent} extend it to multinomial outcomes. We implement both for reference, but each CAVI iteration requires a full $(D{+}1){\times}(D{+}1)$ inversion \emph{per class} --- meaning that the per-sample auxiliary variables break the shared-precision structure. This makes runtime linearly dependent on the number of classes, and orders of magnitude slower than the update for the independent binary probit surrogate \citep{wojnowicz2022}, whose precision is shared across classes and whose inversion can be moved outside the CAVI iterations (Algorithm~\ref{alg:cavi}). The trade-off is that stick-breaking Pólya-Gamma is conjugate for a genuine categorical likelihood, at the cost of an imposed class ordering, whereas the independent binary surrogate is order-invariant but only bounds its categorical likelihood.

\paragraph{Deep Kernel Learning.}
Deep Kernel Learning \citep{wilson2016deep} pairs a deep feature extractor with a Gaussian-Process head; the original work uses KISS-GP / inducing-point approximations to avoid the cubic cost of exact GPs, at the expense of additional approximation and tuning overhead. Our method can be interpreted as a variational approximation to DKL under specific assumptions: marginalizing the last-layer weights under our Gaussian posterior induces a \emph{linear} kernel on the (neural) features, and the independent binary probit surrogate, whose covariance is shared across classes, admits closed-form updates that scale linearly in both $N$ and $K$.

\section{Methods}

The method is a variational filter over the last-layer weights. \S\ref{sec:ibprobit} gives its observation model, \S\ref{sec:cavi} its update step, \S\ref{sec:filtering} its dynamics and prediction step, and \S\ref{sec:moderated} its predictive.

\subsection{Problem Setting}

Consider a classification problem with $K$ classes. Given an image $\bm{i}$ and features $\bx = h(\bm{i}, \pmb{\theta}) \in \R^D$ produced by a neural-network backbone $h$, we seek to model the class-predictive probabilities $p(y = k \mid \bx)$ for $k \in \{1, \ldots, K\}$. In standard practice, this is achieved via a linear layer followed by softmax:
\begin{equation}
    p(y = k \mid \bx, \bw) = \frac{\exp(\bw_k^\top [\bx; 1])}{\sum_{j=1}^K \exp(\bw_j^\top [\bx; 1])},
\end{equation}
where $\bw_k \in \R^{D + 1}$ are the weights projecting features to class $k$. Training proceeds by minimizing the cross-entropy loss with respect to parameters $\bw$, in the case of linear probing, or with respect to both last layer and network parameters $\pmb{\theta}$, in the case of full-network training and fine-tuning.

\subsection{Independent Binary Probit Model}
\label{sec:ibprobit}

Following \citet{wojnowicz2022}, we replace the softmax likelihood with a categorical-from-binary (CB) probit likelihood and perform inference through its independent binary (IB) probit surrogate. We pad the input features with a constant to absorb the bias, writing $\tilde{\bx} = [\bx; 1] \in \R^{D+1}$ and $d = D+1$ for its dimension. The surrogate treats the $K$ entries of the one-hot label vector $\bm z$, with $z_k = \delta_{y, k}$, as independent binary probit observations,
\begin{equation}\label{eq:ib-probit}
    p_{\mathrm{IB}}(\bm{z} \mid \tilde{\bx}, \bm{W}) = \prod_{k=1}^K \Phi(\bw_k^\top \tilde{\bx})^{z_k} \left(1 - \Phi(\bw_k^\top \tilde{\bx})\right)^{1-z_k},
\end{equation}
where $\Phi(\cdot)$ denotes the standard normal cumulative distribution function (CDF). The surrogate is not a categorical model, because it also assigns probability to label vectors with no active class or with several. The categorical likelihood is the CB model obtained by conditioning the surrogate on exactly one active class, which \citet{wojnowicz2022} call the categorical-from-binary-via-conditioning (CBC) model:
\begin{equation}\label{eq:cbc}
    p_{\mathrm{CB}}(y = k \mid \tilde{\bx}, \bm{W}) = \frac{\rho(\bw_k^\top \tilde{\bx})}{\sum_{j=1}^K \rho(\bw_j^\top \tilde{\bx})}, \qquad \rho(a) = \frac{\Phi(a)}{\Phi(-a)}.
\end{equation}
It satisfies $p_{\mathrm{CB}}(y \mid \tilde{\bx}, \bm{W}) > p_{\mathrm{IB}}(\bm{z} \mid \tilde{\bx}, \bm{W})$ for every label, so the evidence lower bound of the surrogate is also a lower bound on the marginal likelihood of the categorical model (Appendix~\ref{app:cb-bound}). We therefore infer the posterior over $\bm W$ with the surrogate (\S\ref{sec:cavi}) and use \eqref{eq:cbc} for the training loss and for prediction (\S\ref{sec:loss-formulations}).

We place a Gaussian prior on the weight matrix $\bm{W} = [\bw_1, \ldots, \bw_K] \in \R^{(D + 1) \times K}$ in its general conjugate form --- a product of multivariate normals whose covariance is shared across classes:
\begin{equation}\label{eq:prior}
    p(\bm{W}) = \prod_{k=1}^K \N(\bw_k \mid \bmu^0_k, \bSigma_0),
    \qquad
    \bLambda_0 = \bSigma_0^{-1}, \quad \bm\eta_{0,k} = \bLambda_0\bmu^0_k .
\end{equation}
At initialisation we take $\bmu^0_k = \bm 0$ and $\bSigma_0 = \bI$, but we keep the prior general because the filtering recursion of \S\ref{sec:filtering} replaces it at every batch with the predicted posterior.

\subsection{The Variational Update: CAVI with Parameter Expansion}
\label{sec:cavi}

\paragraph{Variational posterior.}
We approximate the intractable posterior $p(\bm{W} \mid \mathcal{D})$, where $\mathcal{D} = \{(\bm{i}_n, y_n)\}_{n=1}^N$ denotes a dataset, by the same family as \eqref{eq:prior},
\begin{equation}
    q(\bm{W}) = \prod_{k=1}^K \N(\bw_k \mid \bmu_k, \bSigma), \qquad \bmu_k \in \R^{D+1},\; \bSigma \in \R^{(D+1)\times(D+1)}.
\end{equation}
The shared $\bSigma$ is a consequence of the independent-probit likelihood rather than a particular choice of the variational approximation: the quadratic term in $\bw_k$ is driven by $\tilde\bx_n \tilde\bx_n^\top$ and does not depend on $k$, so every class posterior inherits the same precision and only the first natural parameter differs by class.

\paragraph{Augmented ELBO.}
The independent binary probit admits an Albert--Chib augmentation \citep{albert1993bayesian} with auxiliary variables $\tilde z_{nk} \sim \N(\bw_k^\top \tilde\bx_n, 1)$ and observation rule $z_{nk} = \mathbf{1}[\tilde z_{nk} > 0]$. Under the mean-field factorization $q(\bm W,\tilde{\bm Z}) = q(\bm W) \prod_{n,k} q(\tilde z_{nk})$, the evidence lower bound is
\begin{equation}\label{eq:elbo}
    \mathcal{L}(q) \;=\; \E_q\!\left[\log p(\bm Z \mid \tilde{\bm Z}) + \log p(\tilde{\bm Z} \mid \bm W, \tilde{\bm X}) + \log p(\bm W) - \log q(\bm W,\tilde{\bm Z})\right],
\end{equation}
where $\bm Z = (z_{nk})$ is the one-hot label matrix. We maximize \eqref{eq:elbo} by coordinate ascent over the two factors: $q(\tilde{\bm Z})$ is updated with $q(\bm W)$ held fixed, and then $q(\bm W)$ with $q(\tilde{\bm Z})$ held fixed. For a model with these two factors, coordinate-ascent variational inference (CAVI) and variational Bayesian expectation maximization (VB-EM; \citealp{beal2003variational}) are the same algorithm: the update of $q(\tilde{\bm Z})$ is the E-step and the update of $q(\bm W)$ is the M-step. We call the algorithm CAVI, following \citet{wojnowicz2022}, and its two updates the E-step and the M-step. Both updates are available in closed form for this likelihood (Appendix~\ref{app:cavi}), so no gradient-based optimization of the head is required.

\paragraph{E-step.}
Holding $q(\bm W)$ fixed, $q(\tilde z_{nk})$ is the Albert--Chib truncated normal at $0$ \citep{wojnowicz2022} with mean
\begin{equation}\label{eq:ez}
    \E_q[\tilde z_{nk}] = a_{nk} + z_{nk}\,\frac{\phi(a_{nk})}{\Phi(a_{nk})} - (1 - z_{nk})\,\frac{\phi(a_{nk})}{\Phi(-a_{nk})}, \quad a_{nk} = \bmu_k^\top \tilde\bx_n,
\end{equation}
where $\phi(\cdot)$ is the standard normal PDF and $z_{nk} = \delta_{y_n, k}$. The inverse Mills ratios $\frac{\phi(a)}{\Phi(a)}$ and $\frac{\phi(a)}{\Phi(-a)}$ are evaluated in log-space, $\exp\{\log\phi(a) - \log\Phi(\pm a)\}$, for stability in the tails of $\Phi$. This is a coordinate update, so it uses the plain posterior mean $\bmu_k^\top\tilde\bx_n$; marginalising over $q(\bw_k)$ is a separate operation performed at prediction time (see \S\ref{sec:moderated}).

\paragraph{M-step.}
Holding $q(\tilde{\bm Z})$ fixed, $q(\bw_k)$ is Gaussian with natural parameters
\begin{equation}\label{eq:mstep}
    \bLambda \;=\; \bLambda_0 + \tilde{\bm X}^\top \tilde{\bm X}, \qquad \bm{\eta}_k \;=\; \bm\eta_{0,k} + \tilde{\bm X}^\top \E_q[\tilde{\bm z}_{:k}],
\end{equation}
where $\tilde{\bm X} \in \R^{N \times (D+1)}$ stacks the rows $\tilde\bx_n^\top$ and $\bm{\eta}_k = \bLambda \bmu_k = \bSigma^{-1}\bmu_k$. The variational posterior has exactly the form of the prior \eqref{eq:prior}, which is the closure property \S\ref{sec:filtering} relies on. The precision $\bLambda$ is shared across classes, so all $K$ class-conditional means update in parallel as a single matrix operation and one Cholesky factorization is used for all $K$ classes.

\paragraph{Parameter expansion.}
Plain CAVI alternates \eqref{eq:ez} and \eqref{eq:mstep} and converges slowly along one direction: the overall scale of the class scores. The mean-field factorization removes the correlation between $\bm W$ and $\tilde{\bm Z}$ that is present in the true posterior, so each alternation changes the scale only slightly. Parameter-expanded variational Bayes \citep{qi2007parameter} (hereafter, PX) accelerates convergence in this direction. It adds a scale parameter $\nu > 0$ to the model, which multiplies the prior mean of $\bm W$ by $\nu$, and the prior covariance of $\bm W$ and the variance of the auxiliary variables by $\nu^2$. The expanded model assigns the same probability to the labels for every $\nu$ and is the original model at $\nu = 1$, but the ELBO of a given $q$ depends on $\nu$. Each inner iteration therefore also maximizes the ELBO over $\nu$, which has a closed-form solution, and then returns to the original model by rescaling $\bm W$ and $\tilde{\bm Z}$ by $1/\nu$ (Appendix~\ref{app:px}). Writing $\bm b_k = \tilde{\bm X}^\top\E_q[\tilde{\bm z}_{:k}]$ for the data term of \eqref{eq:mstep}, the resulting update divides this term by $\nu$,
\begin{equation}\label{eq:px-update}
    \bm\eta_k \;\gets\; \bm\eta_{0,k} + \bm b_k/\nu,
    \qquad
    NK\,\nu^2 + c_1\nu - c_2 = 0,
\end{equation}
where $\nu$ is the positive root of the quadratic equation, $N$ is the number of rows of $\tilde{\bm X}$, and $c_1$ and $c_2$ are formed from quantities already computed in the E-step and the M-step. At a fixed point of plain CAVI $c_2 - c_1 = NK$, so $\nu = 1$ and \eqref{eq:px-update} is the ordinary M-step: PX converges to the same solution as plain CAVI in fewer inner iterations. The precision $\bLambda$ is not changed by the expansion, and the inner-iteration count $T$ trades compute for closer convergence to that fixed point.

\subsection{Variational Filtering: the Prior Between Batches}
\label{sec:filtering}

Running CAVI over the full dataset is impractical when the backbone drifts across epochs (full-network fine-tuning) or when $N$ is very large (ImageNet-scale). We therefore process the data in mini-batches $\mathcal{B}_m$ of size $B$. Since the posterior of \S\ref{sec:cavi} has the same form as the prior \eqref{eq:prior}, the same update can be applied to each batch, and it only remains to choose the prior parameters $(\bm\eta_0, \bLambda_0)$ for batch $m$.

We make this choice by treating the head as the latent state of a linear-Gaussian state-space model whose dynamics are a random walk with independent (per-direction) process noise,
\begin{equation}\label{eq:transition}
    \bm{W}_m = \bm{W}_{m-1} + \bXi_m,
    \qquad
    \bXi_m \sim \prod_{k=1}^{K}\N(\bm 0,\, \bQ),
    \qquad
    \bQ = q \bI.
\end{equation}
This is a modelling assumption rather than a numerical stabilisation technique, and it has two independent motivations. The first is drift: under full-network fine-tuning the features $\bx = h(\bm i, \btheta)$ change at every step, so the optimal last layer drifts, and $q$ sets the rate of this drift.

The second is calibration, and it applies even when the backbone is frozen and there is no drift to track. A recursion without process noise accumulates precision without bound, so $\bSigma_m \to \bm 0$ and the moderated predictive of \S\ref{sec:moderated} collapses onto posterior expectation $\bmu$: confidence then grows with the quantity of data the head has seen rather than with the evidence bearing on any particular input, which is the failure mode that a Bayesian treatment is intended to avoid. Process noise instead imposes a lower bound on the posterior covariance, $\Sigma^{*} \approx \sqrt{q/\lambda}$ in the scalar reduction with per-batch data precision $\lambda$ --- the geometric mean of the process noise and the inverse data precision rather than $q$ itself (Appendix~\ref{app:filter}) --- so the moderation term remains non-negligible for arbitrarily long training. Calibration is therefore one of the two motivations for the filtering formulation, and not only a consequence of it.

Propagating the previous posterior through \eqref{eq:transition} increases its covariance and leaves its mean unchanged, so the prior for batch $m$ is
\begin{equation}\label{eq:predict}
    \bSigma^{\mathrm{pred}}_m = \bSigma_{m-1} + \bQ,
    \qquad
    \bLambda^{\mathrm{pred}}_m = \big(\bSigma^{\mathrm{pred}}_m\big)^{-1},
    \qquad
    \bm\eta^{\mathrm{pred}}_{m,k} = \bLambda^{\mathrm{pred}}_m\bmu_{m-1,k},
\end{equation}
and the update is the CAVI/PX step of \S\ref{sec:cavi} run with $(\bLambda_0, \bm\eta_0) \gets (\bLambda^{\mathrm{pred}}_m, \bm\eta^{\mathrm{pred}}_m)$. The prediction step is exact, as in a Kalman filter. The update step is not, because the probit observation is non-Gaussian; it is the variational update of \S\ref{sec:cavi}, which returns a posterior in the family \eqref{eq:prior}. We call the resulting recursion a \emph{variational filter}. Since $\bQ$ does not depend on the class, the precision remains shared across the $K$ classes and the only added cost is the second Cholesky factorization, for $\bSigma^{\mathrm{pred}}_m$.

\paragraph{Setting the process noise.} The process noise $q$ of \eqref{eq:transition} is the one hyperparameter of the head. We parametrize it as
\begin{equation}\label{eq:q-prior}
    q = \frac{\alpha^2}{B},
\end{equation}
where $\alpha$ is the effective learning rate of the head. At the steady state of the filter the posterior mean is an exponentially weighted average of the estimates from the individual mini-batches, in which the newest mini-batch has approximately the weight $\sqrt{\lambda q}$ (Appendix~\ref{app:filter}). For features of unit scale per coordinate $\lambda \approx B$, so this weight is $\alpha$, and a mini-batch influences the posterior mean over approximately $1/\alpha$ filtering steps. The features used in the experiments are not normalised, so there $\alpha$ equals this weight only up to the scale of the features, and $q$ is selected directly by the hyperparameter search (Appendices~\ref{sec:appendix-tuning} and~\ref{sec:ft-search}). The appropriate value depends on the number of training steps $M$: Appendix~\ref{sec:appendix-q} derives the rule $\alpha \propto M^{-1}$, that is, $q \propto M^{-2}$, from the steady state of the filter and reports a measurement that is consistent with it.

\subsection{The Posterior Predictive}
\label{sec:moderated}

For the independent binary probit the required expectation is closed-form (\citealp{bishop2006pattern}; Appendix~\ref{app:moderated}); the relevant measure is the one-step-ahead predictive $\bw_k \sim \N(\bmu_k, \bSigma^{\mathrm{pred}})$ of \eqref{eq:predict}, and because $\bSigma$ and $\bQ$ are both shared across classes the result is a single class-independent rescaling of the class scores:
\begin{equation}\label{eq:moderated}
    \E_{\N(\bmu_k, \bSigma^{\mathrm{pred}})}\big[\Phi(\bw_k^\top\tilde\bx)\big]
    = \Phi\!\left(\frac{\bmu_k^\top\tilde\bx}{s(\tilde\bx)}\right),
    \quad
    s(\tilde\bx) = \sqrt{1 + q\|\tilde\bx\|^2 + \tilde\bx^\top\bSigma\tilde\bx}\,.
\end{equation}
The denominator $s(\tilde\bx)$ is an \emph{input-dependent temperature}: one scalar per input, larger for inputs lying in feature directions the posterior does not constrain, and constant only in the degenerate case --- which is precisely temperature scaling \citep{guo2017calibration}. A predictive evaluated at the posterior mean discards $s$ entirely, so temperature scaling is the constant-$s$ special case of \eqref{eq:moderated}, and the two differ only where $s$ varies across inputs.

The epistemic term $\tilde\bx^\top\bSigma\tilde\bx$ enters the predictive in both regimes. Away from the training distribution it increases, $s$ increases with it, and confidence decreases (see the detection results in \S\ref{sec:results}, which are explained by this mechanism). In distribution it does not vanish, which is the benefit of the filtering formulation: without process noise $\bSigma \to \bm 0$ and $s \to 1$, so the predictive degenerates to the plug-in estimate at $\bmu$ and confidence grows with the amount of data seen. The floor $\Sigma^{*}\approx\sqrt{q/\lambda}$ of \S\ref{sec:filtering} prevents $s$ from collapsing to one as data accumulates; Appendix~\ref{app:filter} derives it, and Appendix~\ref{sec:appendix-tuning} reports the detection result obtained when $q$ is at the lower bound of its search range. The additive term $q\|\tilde\bx\|^2$ is the direct contribution of the process noise to the predictive variance; its size relative to the epistemic term depends on the selected $q$ and on the scale of the features.

\begin{algorithm}[t]
\caption{Variational filtering for the Independent-Binary-Probit last layer}\label{alg:cavi}
\begin{algorithmic}[1]
\Require Features $\{\bx_n = h(\bm i_n)\}$, labels $\{y_n\}$, batch size $B$, inner iterations $T$, process noise $q$ from \eqref{eq:q-prior}
\Ensure $\bm\eta, \bLambda$
\State $\bmu \gets \bm 0$, \ $\bSigma \gets \bI$, \ $\bQ \gets q\bI$
\For{each mini-batch $\mathcal{B}_m$}
    \State $\tilde{\bm X}_m \gets [\,[\bx_n; 1]\,]_{n\in\mathcal{B}_m}$;\quad $\bm Z_m \gets \text{one-hot}(\{y_n\})$
    \State $\bSigma^{\mathrm{pred}}, \bLambda^{\mathrm{pred}}, \bm\eta^{\mathrm{pred}} \gets$ \eqref{eq:predict} \Comment{\textsc{predict}: first Cholesky}
    \State $\bLambda \gets \bLambda^{\mathrm{pred}} + \tilde{\bm X}_m^\top\tilde{\bm X}_m$ \Comment{\eqref{eq:mstep} with $\bLambda_0 = \bLambda^{\mathrm{pred}}$; second Cholesky}
    \State $\bm\eta \gets \bm\eta^{\mathrm{pred}}$
    \For{$t = 0,\dots,T-1$} \Comment{\textsc{update}: PX-CAVI, all $K$ classes in parallel}
        \State $\bm b \gets \tilde{\bm X}_m^\top\,\E_q[\tilde{\bm Z}]$ \Comment{E-step \eqref{eq:ez} at $\bmu = \bLambda^{-1}\bm\eta$}
        \State $\bm\eta \gets \bm\eta^{\mathrm{pred}} + \bm b/\nu$ \Comment{scale $\nu$ from \eqref{eq:px-update}}
    \EndFor
    \State $\bmu \gets \bLambda^{-1}\bm\eta$;\quad $\bSigma \gets \bLambda^{-1}$
\EndFor
\end{algorithmic}
\end{algorithm}

Each mini-batch is one variational filtering step, and every quantity in Algorithm~\ref{alg:cavi} is updated by a closed-form conjugate step: the head has no learning rate, no weight decay, and no gradient-based optimization of any kind. The one free scalar is the process noise $q$ of \eqref{eq:q-prior}. Backbone gradients are taken through \eqref{eq:moderated} alone (Appendix~\ref{sec:ft-procedure}), and the cost of a filtering step is given in Appendix~\ref{app:cost}.

\subsection{Loss Function}
\label{sec:loss-formulations}

The variational filter of \S\ref{sec:filtering} supplies the filtering posterior over the last-layer weights, but training the \emph{backbone} still requires a differentiable scalar loss, and evaluation requires class probabilities. For both we use the categorical likelihood \eqref{eq:cbc}, evaluated at the moderated class scores of \eqref{eq:moderated}. Writing $a_k(\tilde\bx) = \bmu_k^\top\tilde\bx / s(\tilde\bx)$, the predictive class probabilities and the loss are
\begin{equation}\label{eq:cbc-loss}
    p(y = k \mid \tilde\bx) = \frac{\rho\big(a_k(\tilde\bx)\big)}{\sum_{j=1}^K \rho\big(a_j(\tilde\bx)\big)}, \qquad \mathcal{L}_{\text{CBC}} = -\ln p(y \mid \tilde\bx),
\end{equation}
with $\rho(a) = \Phi(a)/\Phi(-a)$ as in \eqref{eq:cbc}. All experiments in this paper use this loss, both for training the backbone and for evaluation.

In the loss and in the predictive class probabilities we replace $\Phi$ by the cubic GELU-style approximation \citep{hendrycks2016gaussian},
\begin{equation}\label{eq:approx-cdf}
    \Phi(x) \;\approx\; \sigma\!\bigl(2\sqrt{2/\pi}\bigl(x + 0.044715\,x^3\bigr)\bigr),
\end{equation}
where $\sigma(\cdot)$ is the logistic function. Under this approximation $\ln\rho(a) = 2\sqrt{2/\pi}\,\bigl(a + 0.044715\,a^3\bigr)$ exactly, so \eqref{eq:cbc-loss} is a softmax cross-entropy on a cubic transformation of the moderated scores. All reported metrics are computed with this approximation; the CAVI inner loop uses the exact $\Phi$ and $\phi$ in log-space.

\section{Results}
\label{sec:results}

Every experiment below uses DINOv3 backbones \citep{simeoni2025dinov3} and
compares three heads under one training protocol: AdamW with cross-entropy and post-hoc
temperature scaling (\textbf{CE+TS}), Variational Bayesian Last Layers
(\textbf{VBLL}, \citealp{harrison2024variational}), and \textbf{IBProbit}, the
variational filter of \S\ref{sec:filtering}. We
report accuracy ($\uparrow$), NLL ($\downarrow$) and Brier score ($\downarrow$).

Note that the cross-entropy baseline is always shown with temperature scaling.
TS is the strongest low-cost calibration baseline: it fits a single parameter post-hoc on the same validation split used for model selection, so it adds only marginal cost to tuning the \textbf{CE+TS} variant. In addition, each training method's hyper-parameters are tuned independently per dataset
under a shared step budget and a single objective
(Appendix~\ref{sec:appendix-tuning}). The backbones, datasets, baselines, metrics and compute resources are described in Appendix~\ref{sec:appendix-setup}. We first study the head in isolation on frozen features
(\S\ref{sec:probing-scale}, \S\ref{sec:ood-detection}), then jointly with the
backbone (\S\ref{sec:fullnetwork}).

\subsection{Linear probing on frozen DINOv3 features}
\label{sec:probing-scale}

\begin{figure}[t!]
    \centering
    \includegraphics[width=\textwidth]{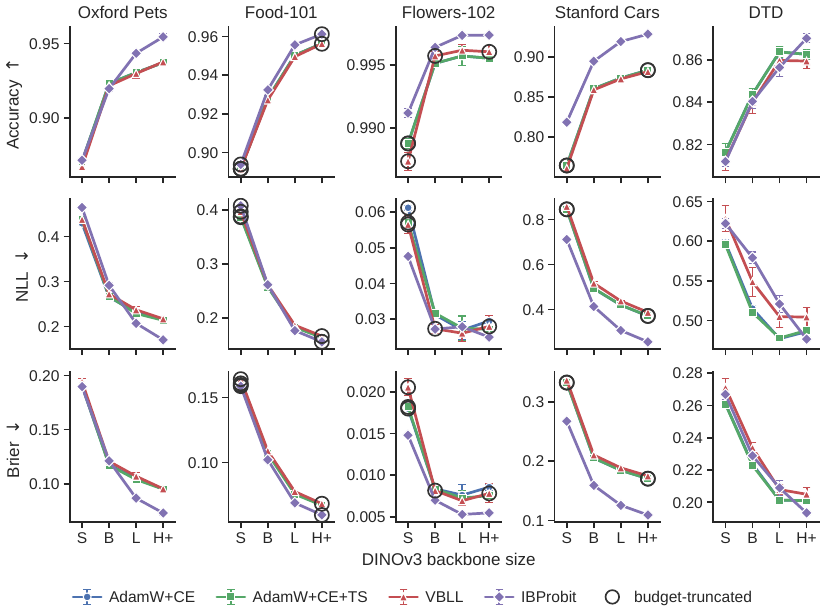}
    \caption{Linear probing on frozen DINOv3 features across backbone scale.
    Columns are the five downstream tasks; rows are accuracy ($\uparrow$), NLL
    ($\downarrow$) and Brier score ($\downarrow$); the $x$-axis is the backbone size (Small, Base, Large, Huge). Points show the mean over ten seeds and
    error bars $\pm 1$ standard deviation. AdamW+CE is shown without and with temperature
    scaling (TS); the temperature is fit on the same validation split used for model selection. A black ring marks a setting in which the validation optimum of at least one seed fell in the last 10\% of the step budget, so that the plotted value is a lower bound (Appendix~\ref{sec:appendix-tuning}). Every method is tuned independently per dataset under one shared 30{,}000-step budget, at DINOv3-Huge only; the other three backbone sizes use the configuration tuned at Huge (Appendix~\ref{sec:appendix-tuning}), so points left of Huge are not separately optimized for any method.}
    \label{fig:probing-scale}
\end{figure}

Frozen features isolate the head: there are no backbone gradients,
and any difference between training methods is attributable to the last layer alone. Every method is trained on the same cached features for the same budget of $30{,}000$ steps and is scored at the evaluation with the lowest validation NLL (Appendix~\ref{sec:appendix-tuning}); for IBProbit a step is one filtering step on a mini-batch, so the filter makes repeated passes over the training data. We
probe four backbone sizes --- DINOv3 Small, Base, Large, Huge --- on five
downstream tasks with training sets of $1.0$K to $75.8$K images and $37$ to $196$ classes, with ten seeds per setting (Figure~\ref{fig:probing-scale}).

At DINOv3-Huge, IBProbit attains the best accuracy and the best NLL (Brier) on all five
tasks. The margin is largest on Stanford Cars, the task with the most classes:
$0.9286$ accuracy against $0.8837$ for CE+TS and $0.8819$ for
VBLL, with NLL $0.2572$ against $0.3723$ and $0.3881$. On Oxford Pets the
accuracy gap is $+1.7$pp over both baselines and NLL falls from $0.214$ (CE+TS) and $0.217$ (VBLL)
to $0.171$. The accuracies of CE+TS and VBLL are within $0.4$pp of each other on every task at this backbone size, and VBLL does not have a lower NLL than CE+TS on any of them. This indicates that a variational last layer trained by gradient
descent improves little on a well-tuned linear probe with a fitted temperature,
whereas a variationally filtered probit head does improve on it.

The number of tasks on which IBProbit leads increases with backbone size. At DINOv3-Small the results are mixed: IBProbit has the lowest NLL on Flowers-102 and
Stanford Cars but trails CE+TS in NLL on Oxford Pets ($0.4639$ vs $0.4366$),
Food-101 ($0.4079$ vs $0.3866$) and DTD ($0.6222$ vs $0.5954$). It also trails CE+TS in NLL on three tasks at DINOv3-Base, on two at DINOv3-Large, and on none at DINOv3-Huge ($29\times$ more parameters than Small). The margin itself does not increase on every task: on Stanford Cars it is largest at DINOv3-Small. We do not attribute the trend with backbone size to the predictive variance. The factor $s(\tilde\bx)$ of \eqref{eq:moderated} rescales all class scores of an input by the same positive number and therefore leaves the predicted class unchanged, so the accuracy differences in Figure~\ref{fig:probing-scale} are differences between the fitted mean weights. We report this trend explicitly because the opposite
would have been the more common result, and because it is relevant to whether
the added complexity of the method is justified at the scales used in practice.

One caveat concerns cost, because \S\ref{sec:fullnetwork} reaches the opposite conclusion at ImageNet scale. IBProbit's uncertainty state is a single covariance shared
across classes --- $d(d+1)/2$ parameters, independent of the class count ---
against $Kd$ for VBLL's per-class diagonal; the two are equal at
$K = (d+1)/2$, which is $193$, $385$, $513$ and $641$ across the four backbone sizes.
Every task here has $K \le 196$, so in nineteen of these twenty settings it is
IBProbit that has the \emph{larger} head: $3.3\times$ VBLL's at Stanford
Cars on Huge, and $17\times$ at Oxford Pets, where a shared $1281 \times 1281$
covariance is compared with $37$ per-class diagonals. The absolute quantities are
negligible either way ($3.1$\,MB at the largest), but the accuracy and NLL
gains above are obtained with a larger head and the
$K$-independence argued in \S\ref{sec:fullnetwork} is a many-class property that
does not apply in the probing regime.

\subsection{Out-of-distribution detection from frozen features}
\label{sec:ood-detection}

In-distribution calibration does not determine behaviour under shift, so we evaluate separately whether probes trained under the same protocol separate in- from
out-of-distribution inputs. We use three pairs at DINOv3-Huge: CIFAR-10 vs SVHN
and CIFAR-100 vs SVHN (far-OOD), and CIFAR-10 vs CIFAR-100 (near-OOD, the harder
of the two CIFAR-10 pairs). CIFAR appears here as \emph{in-distribution} data on a frozen ViT --- chosen because SVHN and the CIFAR-10/100 split provide a clear contrast between far and near OOD on identical features.

Every condition is scored with maximum softmax probability (MSP), and we report AUROC
and FPR at 95\% TPR. For IBProbit this choice does
not discard the uncertainty estimate: the moderated predictive rescales the logits by a class-independent
factor $s(\tilde x)$, which acts as a per-input temperature, so its MSP is a
monotone function of the epistemic term $\tilde x^\top
\bm\Sigma^{\mathrm{pred}}\tilde x$. An MSP advantage under shift is therefore
indirect evidence that the epistemic term is informative.

\begin{figure}[t!]
    \centering
    \includegraphics[width=\textwidth]{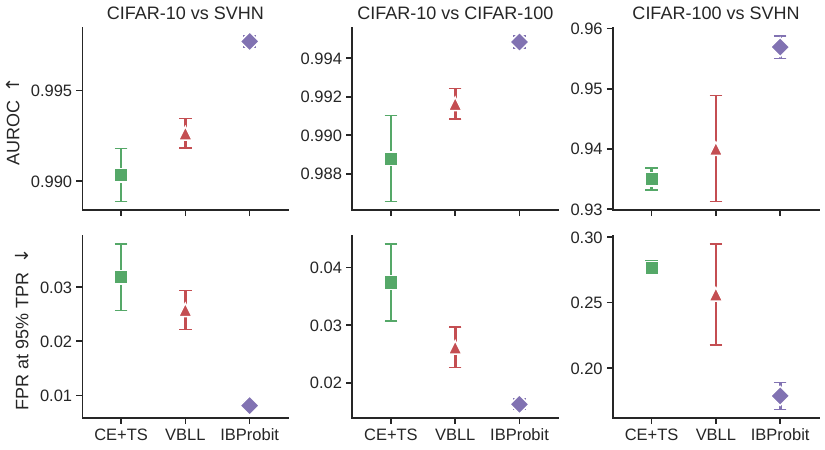}
    \caption{Out-of-distribution detection from frozen DINOv3-Huge features,
    scored by maximum softmax probability. Columns are the three ID/OOD pairs, far-OOD for the two pairs with SVHN and near-OOD for CIFAR-10 vs CIFAR-100;
    rows are AUROC ($\uparrow$) and FPR at 95\% TPR ($\downarrow$). Points show
    the mean over ten seeds with $\pm 1$ standard deviation. AdamW+CE is shown
    with temperature scaling fit on the in-distribution validation split --- the
    same operating point reported in Figure~\ref{fig:probing-scale} --- and the
    single fitted temperature is applied to both the ID and the OOD logits, as
    it would be at deployment.}
    \label{fig:ood-detection}
\end{figure}

IBProbit leads on all three pairs and both metrics
(Figure~\ref{fig:ood-detection}). The largest difference is on CIFAR-10 vs SVHN, where
FPR@95 falls to $0.0081 \pm 0.0007$ against $0.0318 \pm 0.0062$ for CE+TS and
$0.0258 \pm 0.0036$ for VBLL --- a factor of three below the better baseline, on
identical features and an identical evaluation set. On near-OOD the ordering is
preserved with a smaller margin ($0.0163$ against $0.0374$ and $0.0261$), and on
CIFAR-100 vs SVHN, hardest for every method, IBProbit still leads ($0.1788$ against
$0.2763$ and $0.2563$). IBProbit also has the highest in-distribution accuracy on both datasets,
exceeding the next best method by $0.001$ on CIFAR-10 and $0.003$ on CIFAR-100,
so the improved detection does not come at the cost of accuracy. Its
in-distribution NLL is the lowest of the three methods on CIFAR-10 but the
highest on CIFAR-100 ($0.203$, compared with $0.188$ for CE+TS and $0.193$ for
VBLL).

\subsection{Full-network fine-tuning on ImageNet-1k}
\label{sec:fullnetwork}

Probing keeps the features fixed. A more demanding test is whether the filter still helps when the backbone changes during training, which is the regime its process
noise was introduced for. We therefore
fine-tune DINOv3-Huge end to end on ImageNet-1k for two epochs ($2.6$M samples),
five seeds per method, with all three heads under an identical training protocol.
Hyperparameters are tuned at the reported backbone, on the $399{,}872$-sample budget fixed by the tuning protocol ($781$ steps at batch $512$), and transferred to
the reported duration by the AdamW SDE scaling rule \citep{mlodozeniec2025completed}; the process noise of IBProbit is transferred by the rule $q \propto M^{-2}$, where $M$ is the number of training steps. The training step, the search and the transfer are described in Appendix~\ref{sec:appendix-tuning-ft}.

\begin{figure}[t!]
    \centering
    \includegraphics[width=\textwidth]{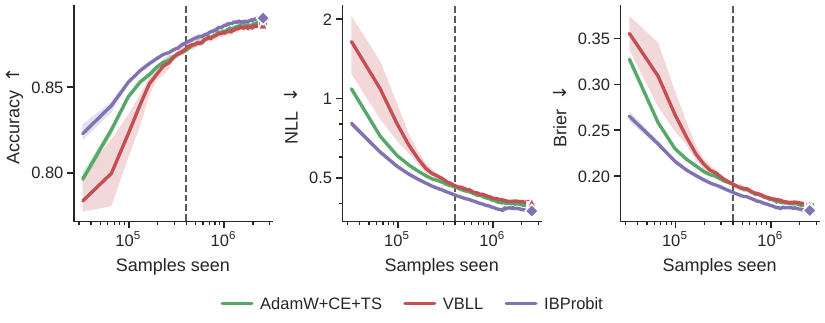}
    \caption{Full-network fine-tuning of DINOv3-Huge on ImageNet-1k: metrics on
    the held-out $80\%$ of the ImageNet validation split against samples seen
    (log scale). Lines show the mean over five seeds, and bands the range
    between the smallest and the largest value across seeds. The dashed vertical line marks the $399{,}872$-sample budget the
    hyperparameters were selected at; results to the right of it lie beyond the budget used for tuning, for every method equally.
    Accuracy, NLL and Brier are shown; NLL and Brier are proper scoring rules
    and support the calibration claim, whereas ECE depends on the choice of binning. Brier is the multiclass definition over the full
    predictive vector. Raw AdamW+CE is omitted: CE+TS is the baseline being
    compared against, and in the accuracy panel the two coincide exactly.}
    \label{fig:sample-efficiency}
\end{figure}

\paragraph{Sample efficiency.} IBProbit has a large advantage early in training
(Figure~\ref{fig:sample-efficiency}). At $66$K samples it reaches $0.839$
accuracy against $0.825$ for CE+TS and $0.800$ for VBLL, with NLL $0.63$ against
$0.72$ and $1.09$; at the first evaluation, $33$K samples, the NLL gap is wider
still --- $0.80$ against $1.09$ and $1.65$. This is the regime the filter is designed for: each batch is assimilated in one closed-form filtering step, so the head is
usable long before a gradient-trained head has converged.

A further result, not available from probing, is that the advantage persists throughout training.
IBProbit stays below both baselines on NLL and Brier and above them on accuracy
for the whole trajectory, and it is still ahead at the endpoint
(Table~\ref{tab:ft-endpoint}), on every metric in distribution and under shift.
Two of the margins are within about one standard deviation, both over VBLL under shift: NLL,
$0.6960 \pm 0.0108$ against $0.7036 \pm 0.0097$, and accuracy, $0.8395 \pm 0.0015$ against $0.8373 \pm 0.0025$. For the other metrics the gap to the better baseline is between two and ten times the larger of the two standard deviations.

\begin{figure}[t!]
    \centering
    \includegraphics[width=0.85\textwidth]{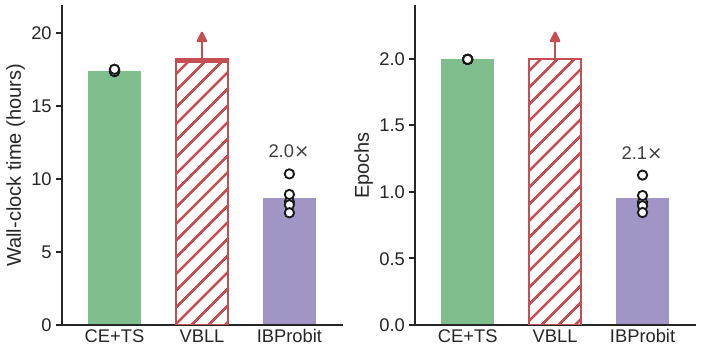}
    \caption{Training needed to reach the final performance of cross-entropy with temperature scaling (CE+TS), in wall-clock hours (left) and in epochs (right). The target is the mean over the five CE+TS seeds of the accuracy, NLL and Brier score at the end of the two-epoch run (Table~\ref{tab:ft-endpoint}). The target is \emph{joint}: a run reaches it at the first evaluation at which all three metrics are at or better than the target. The CE+TS bar is the length of its own runs. For IBProbit the bar shows the mean over the five seeds and circles the individual seeds; all five reach the target, and the number above the bar is the speed-up relative to CE+TS. No VBLL run reaches the target within two epochs, so VBLL is right-censored: the hatched bar with an arrow marks the length of its runs, which is a lower bound. Time is measured from the first evaluation of each run. The times of CE+TS and IBProbit include the evaluation of the shift sets during training, which the VBLL runs delegate to a separate process.}
    \label{fig:cost-to-threshold}
\end{figure}

\paragraph{Cost to reach a target performance.} Fixing a target performance and measuring the
cost to reach it makes the comparison concrete
(Figure~\ref{fig:cost-to-threshold}). We take as the target the performance that cross-entropy with temperature scaling, the baseline with the better in-distribution results, has at the end of its two-epoch run: the mean over seeds of its accuracy, NLL and Brier score at the final evaluation, which are the values in Table~\ref{tab:ft-endpoint}. For the other two methods we measure the amount of training after which a run first meets all three.

Every IBProbit run reaches the target. On average it does so after $8.7$ hours and $0.95$ epochs, compared with the $17.4$ hours and two epochs for which the cross-entropy runs were trained. This is a speed-up of $2.0\times$ in wall-clock time and $2.1\times$ in samples; across the five seeds the speed-up in wall-clock time ranges from $1.7\times$ to $2.3\times$. No VBLL run reaches the target within its two epochs, which took $18.1$ hours, so VBLL is right-censored in the figure. The lowest NLL of any VBLL run at any evaluation is $0.399$, compared with the target of $0.389$.

The comparison cannot be made in the opposite direction, because neither baseline reaches the performance of IBProbit. Neither the five cross-entropy runs nor the five VBLL runs reach, within the full two-epoch budget, what IBProbit attains after one epoch, jointly on accuracy, NLL and Brier. The shortfall is in calibration rather than in accuracy: every
cross-entropy run passes the accuracy that IBProbit has after one epoch at some point in its trajectory, but
none reaches an NLL below $0.3872$ or a Brier score below $0.1659$, compared with $0.3758$ and $0.1648$ for IBProbit after one epoch.

We do not report peak training memory. At this backbone size the methods' activations dominate and the measurement cannot resolve the head: the state stored by the two Bayesian heads differs by under $2$\,MB, a few thousandths of a percent of the
peak, and the allocator statistics available in the two frameworks are not
defined identically. The head-state comparison below is analytical for that
reason.

In addition to the $K \times d$ mean of any linear head, IBProbit adds
one covariance shared across classes --- $d(d+1)/2$ parameters, \emph{independent
of $K$} --- while VBLL in its diagonal parameterization adds a per-class
diagonal, $Kd$, growing linearly in the class count. At DINOv3-Huge
($d = 1281$) the two are equal at $K \approx 641$: on ImageNet-1k IBProbit's head has $0.82$M uncertainty parameters against VBLL's $1.28$M, $36\%$ fewer,
while on the small-$K$ probing tasks the ordering reverses ($0.82$M against
$0.25$M at Stanford Cars' $K = 196$; see \S\ref{sec:probing-scale}). Richer VBLL parameterizations only widen
the gap in the many-class regime: low rank costs $K d r$ and full covariance
$K d^2$, both of which retain the linear-in-$K$ scaling that the shared
covariance removes.

\begin{figure}[t!]
    \centering
    \includegraphics[width=\textwidth]{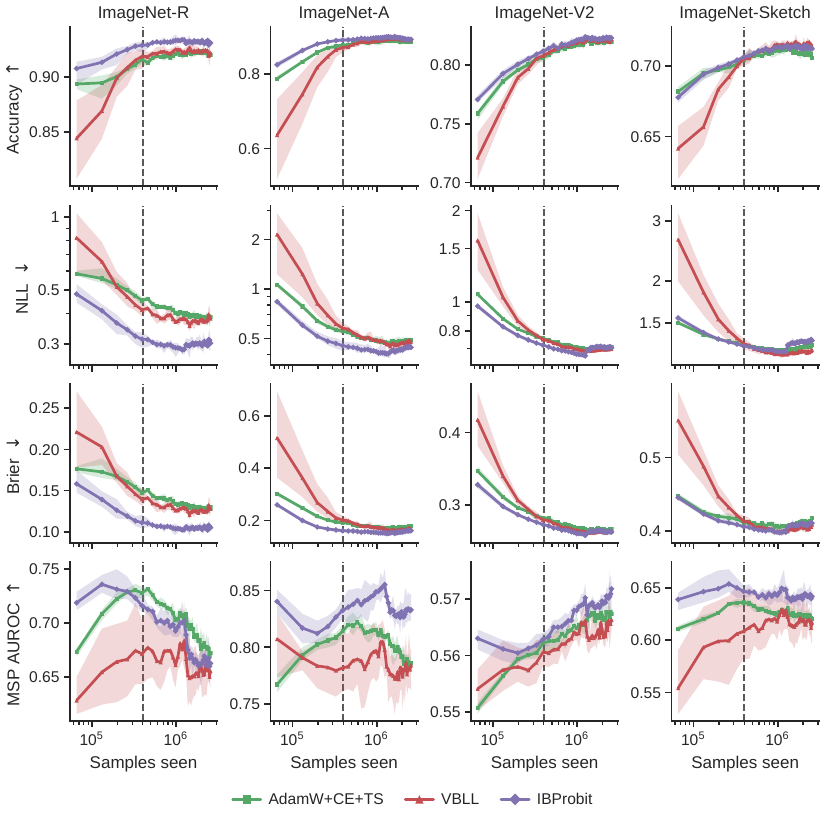}
    \caption{The same fine-tuning trajectories scored under distribution shift.
    Columns are the four shift sets; rows are accuracy, NLL, Brier score and the
    AUROC of maximum-softmax-probability detection against the in-distribution
    split. Lines, bands and the dashed vertical line are as in
    Figure~\ref{fig:sample-efficiency}, and markers show the individual evaluations.
    ImageNet-R and -A cover 200 of the 1000 classes and are evaluated with a
    subset mask; ImageNet-V2 and -Sketch span all 1000 and are not masked. Every method is fit in distribution and then evaluated under shift; in particular the single temperature of the CE baseline is fitted on the in-distribution validation
    split and applied unchanged to the shifted logits, never refitted on the
    shifted set. The shift sets are scored at every second evaluation of the trajectory, so these panels contain fewer points than Figure~\ref{fig:sample-efficiency}.}
    \label{fig:shift}
\end{figure}

\paragraph{Behaviour under shift.} Scoring the same trajectory on ImageNet-R,
-A, -V2 and -Sketch reproduces the ordering on three of the four sets (Figure~\ref{fig:shift}). On ImageNet-R and -A IBProbit has the best accuracy, NLL and Brier score for almost the entire run. On ImageNet-V2 it has the best accuracy throughout and the best NLL in the first half of the run, after which the three methods are close. On ImageNet-Sketch the three methods are close in accuracy and Brier score, and IBProbit has the highest NLL at the endpoint ($1.33$, against $1.29$ for CE+TS and $1.25$ for VBLL). Table~\ref{tab:ft-endpoint}
gives the endpoint averaged over the four sets.

The fourth row scores the same predictives as a \emph{detector}, by maximum
softmax probability, against the in-distribution split at the same step --- the
score used in \S\ref{sec:ood-detection}, so that the two settings are evaluated with the same score. Because ImageNet-R and -A cover $200$ of the $1000$ classes, the
in-distribution reference for those columns is restricted to the same classes
and to the images whose label falls in them; comparing a $200$-class maximum
against a $1000$-class one would measure the class count rather than the shift.
IBProbit detects best at the first evaluation on all four sets, which is consistent with the sample-efficiency result, and it still leads at the
endpoint on three of the four; cross-entropy overtakes it only on ImageNet-R
($0.673$ against $0.663$). On ImageNet-V2 no method detects the shift much above chance, which is expected for a re-collection of the same classes.

One feature of this row requires an explanation.
On ImageNet-R, and to a smaller extent on ImageNet-A and -Sketch, detection quality peaks partway through training and then falls for
\emph{every} method --- on ImageNet-R cross-entropy loses $0.059$ AUROC from its own
peak, IBProbit $0.073$, VBLL $0.028$ --- while the accuracy, NLL and Brier rows
above are stable or improving over the same period. Detection and calibration therefore diverge under continued fine-tuning. This is a consequence of fitting the in-distribution task more closely and is not specific to any head: a model that becomes more confident on ImageNet also becomes more confident on ImageNet-R, and cross-entropy, which has no posterior covariance, shows the effect as strongly as the Bayesian methods. We therefore do not interpret it as evidence about the covariance floor of \S\ref{sec:filtering}.

\begin{table}[ht!]
\centering
\caption{Fine-tuning endpoint at $2.56$M samples, the largest evaluation present
in every run. Mean $\pm$ standard deviation over five seeds. The rows under shift
average ImageNet-R, -A, -V2 and -Sketch; AUROC is maximum-softmax-probability
detection against the in-distribution split at the same step, with the
in-distribution reference restricted to the covered classes on the two masked
sets. ECE is omitted for the reason given in
Figure~\ref{fig:sample-efficiency}'s caption; Brier is the multiclass
definition over the full predictive vector.}
\label{tab:ft-endpoint}
\small
\begin{tabular}{lccc}
\toprule
Metric & AdamW+CE+TS & VBLL & IBProbit \\
\midrule
\multicolumn{4}{l}{\textit{In distribution}} \\
Accuracy $\uparrow$ & 0.8880 $\pm$ 0.0011 & 0.8866 $\pm$ 0.0004 & $\bm{0.8904 \pm 0.0007}$ \\
NLL $\downarrow$ & 0.3891 $\pm$ 0.0015 & 0.4034 $\pm$ 0.0010 & $\bm{0.3744 \pm 0.0010}$ \\
Brier $\downarrow$ & 0.1669 $\pm$ 0.0008 & 0.1693 $\pm$ 0.0006 & $\bm{0.1628 \pm 0.0003}$ \\
\midrule
\multicolumn{4}{l}{\textit{Under shift (mean of four sets)}} \\
Accuracy $\uparrow$ & 0.8323 $\pm$ 0.0017 & 0.8373 $\pm$ 0.0025 & $\bm{0.8395 \pm 0.0015}$ \\
NLL $\downarrow$ & 0.7194 $\pm$ 0.0074 & 0.7036 $\pm$ 0.0097 & $\bm{0.6960 \pm 0.0108}$ \\
Brier $\downarrow$ & 0.2478 $\pm$ 0.0024 & 0.2422 $\pm$ 0.0026 & $\bm{0.2349 \pm 0.0024}$ \\
AUROC $\uparrow$ & 0.6617 $\pm$ 0.0023 & 0.6557 $\pm$ 0.0104 & $\bm{0.6774 \pm 0.0040}$ \\
\bottomrule
\end{tabular}
\end{table}

\section{Limitations}
\label{sec:limitations}

\paragraph{One hyperparameter does not transfer across training durations.} The
process noise $q$ has to be rescaled when the training duration changes; if it is left unchanged, it degrades the head late in training, and this can be mistaken for a property of the method. We give a rule for it in
Appendix~\ref{sec:appendix-q}, derived from the steady state of the filter, but its exponent is measured on one backbone and we
have not established it across widths. A search run at a short budget
and scored by the best value anywhere on its curve will also prefer configurations whose validation loss is low early and then degrades, for the same reason; our fine-tuning searches score the end of the trajectory instead. Both are properties of tuning any method
with a forgetting rate against a truncated budget rather than of this head
alone, but we encountered both in our experiments and therefore report them.

\paragraph{The head's memory advantage is structural, not measurable
end-to-end.} IBProbit's uncertainty state is $K$-independent and smaller than
VBLL's diagonal once $K > (d+1)/2$, but at $d = 1281$ that advantage is under
$2$\,MB, which no end-to-end training measurement can resolve: backbone
activations dominate the peak at this backbone size, and the two frameworks'
allocator statistics are not defined identically, so we do not report peak
memory at all. Isolating the head's footprint would need a measurement designed
for it, which we have not made. The $K$-independence claim is also stated at the
diagonal parameterization we ran; against full-covariance VBLL the gap widens,
but we did not train that variant and do not report it.

\paragraph{Model assumptions.} The posterior covariance shared across classes is
what makes the CAVI inner loop linear in the class count, but it offers no
per-class adaptive capacity, so severe class imbalance may under-regularize
minority classes. The class-independence assumption inherited from
\citet{wojnowicz2022} did not degrade calibration on any task we evaluated, but may be
more restrictive under strong inter-class structure such as hierarchical labels.

\paragraph{Scope of the tuning.} Probing configurations were searched at
DINOv3-Huge only and applied unchanged at the three smaller backbones, whose features have different dimension; the fine-tuning configurations were searched
at DINOv3-Huge on a budget of $0.4$M samples, about a third of an epoch, and transferred to two epochs (Appendix~\ref{sec:appendix-tuning-ft}). Both limitations apply
symmetrically to all three methods, so they affect the absolute numbers rather than the comparison, but the points for the smaller backbones in Figure~\ref{fig:probing-scale} should not be interpreted as the best achievable result of any method.

\paragraph{What the detection result does and does not show.} Figure
\ref{fig:ood-detection} scores MSP, which for IBProbit is a monotone function of
the epistemic term rather than a direct measurement of it. The result is therefore
evidence that the term is informative, not a measurement of its magnitude. The probing runs also store a single evaluation each, so \S\ref{sec:probing-scale} and \S\ref{sec:ood-detection} contain no trajectories: no
detection-versus-compute curve and no endpoint-penalty analysis are available
there, only at fine-tuning scale.

\paragraph{Calibration is not safety.} Better-calibrated probabilities reduce
undetected over-confidence where downstream actions depend on probability thresholds, but they do not address fairness, bias, or adversarial robustness,
and a well-calibrated classifier can still be systematically wrong on
populations under-represented at training time.

\section{Conclusion}

We presented a variational filtering algorithm for the Bayesian last layer, built on the independent binary probit (IBProbit) surrogate of \citet{wojnowicz2022}. By replacing the softmax with a categorical-from-binary probit likelihood and performing inference through its independent binary probit surrogate, we derived efficient closed-form variational filtering updates. This distinguishes our method from recent approaches like VBLL, which keep the softmax, bound it, and rely on gradient-based optimization. Our experiments on DINOv3 backbones demonstrate that an IBProbit based last layer provides a computationally efficient way to obtain well-calibrated deep networks. The closed-form nature of our updates makes this approach particularly attractive for one-shot adaptation and scenarios where gradient-based tuning of the uncertainty mechanism is undesirable. In practice, the method can replace a standard linear probe without further changes to the training pipeline: the head has no learning rate or weight decay and needs no post-hoc calibration step. With the largest backbone it gave higher accuracy and lower negative log-likelihood than both baselines on all five probing tasks, and under full-network fine-tuning on ImageNet-1k it remained better than both on accuracy, negative log-likelihood and Brier score after two epochs. Its uncertainty state is a single covariance matrix shared across classes, so its size does not grow with the number of classes.

Better-calibrated probabilities reduce the risk of undetected over-confidence in high-stakes applications where downstream actions depend on probability thresholds. However, they do not address fairness, bias, or adversarial robustness, and should not be interpreted as a safety certification --- a well-calibrated classifier can still be systematically wrong on populations under-represented at training time.

A natural extension of the presented approach would be a fully end-to-end Bayesian treatment of deep learning that combines the IBProbit head with the IVON optimizer \citep{shen2024variational}, or the more recent EVON optimizer \citep{minut2026soap} for variational inference on the backbone weights.

\section*{Acknowledgments}
The authors gratefully acknowledge the computing time granted on the
high-performance computers Raven (NVIDIA A100) and Viper (AMD Instinct MI300A)
at the Max Planck Computing and Data Facility (MPCDF), and thank the MPCDF
support staff for their assistance.

\bibliographystyle{plainnat}
\bibliography{references}

@article{minut2026soap,
  title={SOAP-Bubbles: Structured Weight Uncertainty for Neural Networks},
  author={Minut, Adrian Robert and Daheim, Nico and Miani, Marco and Khan, Mohammad Emtiyaz and Lin, Wu and M{\"o}llenhoff, Thomas},
  journal={arXiv preprint arXiv:2606.23357},
  year={2026}
}

@inproceedings{wojnowicz2022,
  title={Easy Variational Inference for Categorical Models via an Independent Binary Approximation},
  author={Wojnowicz, Michael and Aeron, Shuchin and Miller, Eric and Hughes, Michael C.},
  booktitle={Proceedings of the 39th International Conference on Machine Learning},
  pages={23879--23899},
  year={2022},
  volume={162},
  series={Proceedings of Machine Learning Research},
  publisher={PMLR}
}

@InProceedings{guo2017calibration,
  title = 	 {On Calibration of Modern Neural Networks},
  author =       {Chuan Guo and Geoff Pleiss and Yu Sun and Kilian Q. Weinberger},
  booktitle = 	 {Proceedings of the 34th International Conference on Machine Learning},
  pages = 	 {1321--1330},
  year = 	 {2017},
  editor = 	 {Precup, Doina and Teh, Yee Whye},
  volume = 	 {70},
  series = 	 {Proceedings of Machine Learning Research},
  month = 	 {06--11 Aug},
  publisher =    {PMLR},
  url = 	 {https://proceedings.mlr.press/v70/guo17a.html}
}

@inproceedings{daxberger2021laplace,
  title={Laplace Redux -- Effortless {Bayesian} Deep Learning},
  author={Daxberger, Erik and Kristiadi, Agustinus and Immer, Alexander and Eschenhagen, Runa and Bauer, Matthias and Hennig, Philipp},
  booktitle={Advances in Neural Information Processing Systems},
  year={2021}
}

@inproceedings{kristiadi2020being,
  title={Being {Bayesian}, even just a bit, fixes overconfidence in {ReLU} networks},
  author={Kristiadi, Agustinus and Hein, Matthias and Hennig, Philipp},
  booktitle={International Conference on Machine Learning},
  pages={5436--5446},
  year={2020},
  organization={PMLR}
}

@InProceedings{blundell2015weight,
  title = 	 {Weight Uncertainty in Neural Network},
  author = 	 {Blundell, Charles and Cornebise, Julien and Kavukcuoglu, Koray and Wierstra, Daan},
  booktitle = 	 {Proceedings of the 32nd International Conference on Machine Learning},
  pages = 	 {1613--1622},
  year = 	 {2015},
  editor = 	 {Bach, Francis and Blei, David},
  volume = 	 {37},
  series = 	 {Proceedings of Machine Learning Research},
  address = 	 {Lille, France},
  month = 	 {07--09 Jul},
  publisher =    {PMLR},
  url = 	 {https://proceedings.mlr.press/v37/blundell15.html}
}

@article{snoek2019can,
  title={Can you trust your model's uncertainty? {Evaluating} predictive uncertainty under dataset shift},
  author={Snoek, Jasper and Ovadia, Yaniv and Fertig, Emily and Lakshminarayanan, Balaji and Nowozin, Sebastian and Sculley, D. and Dillon, Joshua and Ren, Jie and Nado, Zachary},
  journal={Advances in Neural Information Processing Systems},
  volume={32},
  year={2019}
}

@InProceedings{izmailov2021bayesian,
  title = 	 {What Are Bayesian Neural Network Posteriors Really Like?},
  author =       {Izmailov, Pavel and Vikram, Sharad and Hoffman, Matthew D and Wilson, Andrew Gordon Gordon},
  booktitle = 	 {Proceedings of the 38th International Conference on Machine Learning},
  pages = 	 {4629--4640},
  year = 	 {2021},
  editor = 	 {Meila, Marina and Zhang, Tong},
  volume = 	 {139},
  series = 	 {Proceedings of Machine Learning Research},
  month = 	 {18--24 Jul},
  publisher =    {PMLR},
  url = 	 {https://proceedings.mlr.press/v139/izmailov21a.html}
}

@article{albert1993bayesian,
  title={Bayesian analysis of binary and polychotomous response data},
  author={Albert, James H. and Chib, Siddhartha},
  journal={Journal of the American Statistical Association},
  volume={88},
  number={422},
  pages={669--679},
  year={1993},
  publisher={Taylor \& Francis}
}

@inproceedings{kingma2014adam,
  title={Adam: A method for stochastic optimization},
  author={Kingma, Diederik P. and Ba, Jimmy},
  booktitle={International Conference on Learning Representations},
  year={2015}
}

@article{kidger2021equinox,
  title={Equinox: neural networks in {JAX} via callable {PyTrees} and filtered transformations},
  author={Kidger, Patrick and Garcia, Cristian},
  journal={arXiv preprint arXiv:2111.00254},
  year={2021}
}

@software{jax2018github,
  author = {Bradbury, James and Frostig, Roy and Hawkins, Peter and Johnson, Matthew James and Leary, Chris and Maclaurin, Dougal and Necula, George and Paszke, Adam and VanderPlas, Jake and Wanderman-Milne, Skye and Zhang, Qiao},
  title = {{JAX}: composable transformations of {Python}+{NumPy} programs},
  url = {http://github.com/google/jax},
  version = {0.3.13},
  year = {2018}
}

@book{bishop2006pattern,
  title={Pattern Recognition and Machine Learning},
  author={Bishop, Christopher M.},
  publisher={Springer},
  year={2006}
}

@inproceedings{harrison2024variational,
  title={Variational {Bayesian} Last Layers},
  author={Harrison, James and Willes, John and Snoek, Jasper},
  booktitle={International Conference on Learning Representations},
  year={2024}
}

@article{fiedler2023improved,
  title={Improved Uncertainty Quantification for Neural Networks with {Bayesian} Last Layer},
  author={Fiedler, Felix and Lucia, Sergio},
  journal={IEEE Access},
  year={2023},
  note={arXiv:2302.10975}
}

@inproceedings{mukhoti2020calibrating,
  title={Calibrating Deep Neural Networks using Focal Loss},
  author={Mukhoti, Jishnu and Kulharia, Viveka and Sanyal, Amartya and Golodetz, Stuart and Torr, Philip H. S. and Dokania, Puneet K.},
  booktitle={Advances in Neural Information Processing Systems},
  year={2020}
}

@inproceedings{thulasidasan2019mixup,
  title={On Mixup Training: Improved Calibration and Predictive Uncertainty for Deep Neural Networks},
  author={Thulasidasan, Sunil and Chennupati, Gopinath and Bilmes, Jeff A. and Bhattacharya, Tanmoy and Michalak, Sarah},
  booktitle={Advances in Neural Information Processing Systems},
  year={2019}
}

@inproceedings{minderer2021revisiting,
  title={Revisiting the Calibration of Modern Neural Networks},
  author={Minderer, Matthias and Djolonga, Josip and Romijnders, Rob and Hubis, Frances and Zhai, Xiaohua and Houlsby, Neil and Tran, Dustin and Lucic, Mario},
  booktitle={Advances in Neural Information Processing Systems},
  year={2021}
}

@inproceedings{kristiadi2021infinite,
  title={An Infinite-Feature Extension for {Bayesian} {ReLU} Nets That Fixes Their Asymptotic Overconfidence},
  author={Kristiadi, Agustinus and Hein, Matthias and Hennig, Philipp},
  booktitle={Advances in Neural Information Processing Systems},
  year={2021}
}

@inproceedings{lakshminarayanan2017simple,
  title={Simple and Scalable Predictive Uncertainty Estimation using Deep Ensembles},
  author={Lakshminarayanan, Balaji and Pritzel, Alexander and Blundell, Charles},
  booktitle={Advances in Neural Information Processing Systems},
  year={2017}
}

@inproceedings{gal2016dropout,
  title={Dropout as a {Bayesian} Approximation: Representing Model Uncertainty in Deep Learning},
  author={Gal, Yarin and Ghahramani, Zoubin},
  booktitle={International Conference on Machine Learning},
  year={2016}
}

@inproceedings{maddox2019simple,
  title={A Simple Baseline for {Bayesian} Uncertainty in Deep Learning},
  author={Maddox, Wesley J. and Izmailov, Pavel and Garipov, Timur and Vetrov, Dmitry P. and Wilson, Andrew Gordon},
  booktitle={Advances in Neural Information Processing Systems},
  year={2019}
}

@article{polson2013bayesian,
  title={Bayesian Inference for Logistic Models Using {P\'olya}--{Gamma} Latent Variables},
  author={Polson, Nicholas G. and Scott, James G. and Windle, Jesse},
  journal={Journal of the American Statistical Association},
  volume={108},
  number={504},
  pages={1339--1349},
  year={2013}
}

@inproceedings{linderman2015dependent,
  title={Dependent Multinomial Models Made Easy: Stick-Breaking with the {P\'olya}-{Gamma} Augmentation},
  author={Linderman, Scott and Johnson, Matthew J. and Adams, Ryan P.},
  booktitle={Advances in Neural Information Processing Systems},
  year={2015}
}

@inproceedings{wilson2016deep,
  title={Deep Kernel Learning},
  author={Wilson, Andrew Gordon and Hu, Zhiting and Salakhutdinov, Ruslan and Xing, Eric P.},
  booktitle={Artificial Intelligence and Statistics},
  pages={370--378},
  year={2016},
  organization={PMLR}
}

@article{hendrycks2016gaussian,
  title={Gaussian error linear units ({GELU}s)},
  author={Hendrycks, Dan and Gimpel, Kevin},
  journal={arXiv preprint arXiv:1606.08415},
  year={2016}
}

@inproceedings{shen2024variational,
  title={Variational learning is effective for large deep networks},
  author={Shen, Yuesong and Daheim, Nico and Cong, Bai and Nickl, Peter and Marconi, Gian Maria and Bazan, Cl{\'e}ment and Yokota, Rio and Gurevych, Iryna and Cremers, Daniel and Khan, Mohammad Emtiyaz and M{\"o}llenhoff, Thomas},
  booktitle={Proceedings of the 41st International Conference on Machine Learning},
  year={2024},
  series={Proceedings of Machine Learning Research},
  publisher={PMLR}
}

@phdthesis{beal2003variational,
  title={Variational algorithms for approximate Bayesian inference},
  author={Beal, Matthew J.},
  year={2003},
  school={Gatsby Computational Neuroscience Unit, University College London}
}

@inproceedings{broderick2013streaming,
  title={Streaming Variational {Bayes}},
  author={Broderick, Tamara and Boyd, Nicholas and Wibisono, Andre and Wilson, Ashia C. and Jordan, Michael I.},
  booktitle={Advances in Neural Information Processing Systems},
  year={2013}
}

@inproceedings{titsias2024kalman,
  title={{Kalman} Filter for Online Classification of Non-Stationary Data},
  author={Titsias, Michalis K. and Galashov, Alexandre and Rannen-Triki, Amal and Pascanu, Razvan and Teh, Yee Whye and Bornschein, Jorg},
  booktitle={International Conference on Learning Representations},
  year={2024}
}

@inproceedings{qi2007parameter,
  title={Parameter expanded variational {B}ayesian methods},
  author={Qi, Yuan and Jaakkola, Tommi S.},
  booktitle={Advances in Neural Information Processing Systems},
  year={2007}
}

@inproceedings{akiba2019optuna,
  title={Optuna: A Next-generation Hyperparameter Optimization Framework},
  author={Akiba, Takuya and Sano, Shotaro and Yanase, Toshihiko and Ohta, Takeru and Koyama, Masanori},
  booktitle={Proceedings of the 25th ACM SIGKDD International Conference on Knowledge Discovery and Data Mining},
  pages={2623--2631},
  year={2019}
}

@article{simeoni2025dinov3,
  title={DINOv3},
  author={Sim{\'e}oni, Oriane and Vo, Huy V. and Seitzer, Maximilian and Baldassarre, Federico and Oquab, Maxime and others},
  journal={arXiv preprint arXiv:2508.10104},
  year={2025}
}

@inproceedings{deng2009imagenet,
  title={ImageNet: A Large-Scale Hierarchical Image Database},
  author={Deng, Jia and Dong, Wei and Socher, Richard and Li, Li-Jia and Li, Kai and Fei-Fei, Li},
  booktitle={IEEE Conference on Computer Vision and Pattern Recognition},
  year={2009}
}

@inproceedings{loshchilov2019decoupled,
  title={Decoupled Weight Decay Regularization},
  author={Loshchilov, Ilya and Hutter, Frank},
  booktitle={International Conference on Learning Representations},
  year={2019}
}

@article{mlodozeniec2025completed,
  title={Completed Hyperparameter Transfer across Modules, Width, Depth, Batch and Duration},
  author={Mlodozeniec, Bruno and Ablin, Pierre and B{\'e}thune, Louis and Busbridge, Dan and Klein, Michal and Ramapuram, Jason and Cuturi, Marco},
  journal={arXiv preprint arXiv:2512.22382},
  year={2025}
}

\appendix

\section{Derivations}
\label{app:derivations}

This appendix derives the results that \S\ref{sec:ibprobit}--\S\ref{sec:moderated} state without proof: the lower bound of \S\ref{sec:ibprobit} (Appendix~\ref{app:cb-bound}), the two updates of \S\ref{sec:cavi} (Appendix~\ref{app:cavi}), the parameter expansion (Appendix~\ref{app:px}), the steady state of the filter of \S\ref{sec:filtering} (Appendix~\ref{app:filter}), the predictive of \S\ref{sec:moderated} (Appendix~\ref{app:moderated}) and the cost of a filtering step (Appendix~\ref{app:cost}). Throughout, $d = D+1$ is the dimension of the padded feature $\tilde\bx$, the matrix $\tilde{\bm X} \in \R^{N\times d}$ contains the features of the data being assimilated, which in the filter is one mini-batch ($N = B$), and $a_{nk} = \bmu_k^\top\tilde\bx_n$ are the class scores at the posterior mean.

\subsection{The categorical likelihood and its lower bound}
\label{app:cb-bound}

Let $a_k = \bw_k^\top\tilde\bx$ and let $\bm e_k$ denote the one-hot vector of class $k$. Because $1-\Phi(a) = \Phi(-a)$ and $\rho(a) = \Phi(a)/\Phi(-a)$, the surrogate \eqref{eq:ib-probit} assigns to this vector the probability
\begin{equation}\label{eq:ib-onehot}
    p_{\mathrm{IB}}(\bm e_k \mid \tilde\bx, \bm W) \;=\; \Phi(a_k)\prod_{j\neq k}\Phi(-a_j) \;=\; \rho(a_k)\prod_{j=1}^{K}\Phi(-a_j).
\end{equation}
The probability that exactly one class is active is therefore $P_1 = \sum_{j} p_{\mathrm{IB}}(\bm e_j \mid \tilde\bx, \bm W) = \prod_{i}\Phi(-a_i)\,\sum_{j}\rho(a_j)$, and conditioning the surrogate on this event gives
\begin{equation}
    p_{\mathrm{CB}}(y = k \mid \tilde\bx, \bm W) \;=\; \frac{p_{\mathrm{IB}}(\bm e_k \mid \tilde\bx, \bm W)}{P_1} \;=\; \frac{\rho(a_k)}{\sum_{j}\rho(a_j)},
\end{equation}
which is \eqref{eq:cbc}. Since $P_1 < 1$, we have $p_{\mathrm{CB}}(y = k \mid \tilde\bx, \bm W) > p_{\mathrm{IB}}(\bm e_k \mid \tilde\bx, \bm W)$ for every class and every $\bm W$. The inequality is preserved by the product over the data and by the expectation over the prior $p(\bm W)$. The marginal likelihood of the categorical model is therefore larger than that of the surrogate, and the ELBO \eqref{eq:elbo} of the surrogate is a lower bound on both.

\subsection{The E-step and the M-step}
\label{app:cavi}

\paragraph{Augmented model.} With the auxiliary variables of \S\ref{sec:cavi}, the joint distribution of the labels, the auxiliary variables and the weights is
\begin{equation}\label{eq:joint}
    p(\bm Z, \tilde{\bm Z}, \bm W) \;=\; p(\bm Z \mid \tilde{\bm Z})\,\prod_{n,k}\N\big(\tilde z_{nk} \mid \bw_k^\top\tilde\bx_n,\, 1\big)\,\prod_{k}\N\big(\bw_k \mid \bmu^0_k,\, \bSigma_0\big),
\end{equation}
where $p(\bm Z \mid \tilde{\bm Z})$ is one if every $\tilde z_{nk}$ has the sign prescribed by its label ($\tilde z_{nk} > 0$ for $z_{nk} = 1$ and $\tilde z_{nk} \le 0$ for $z_{nk} = 0$) and zero otherwise. Integrating out $\tilde z_{nk}$ returns the surrogate \eqref{eq:ib-probit}, because $\Pr(\tilde z_{nk} > 0 \mid \bw_k) = \Phi(\bw_k^\top\tilde\bx_n)$. Coordinate ascent on the ELBO \eqref{eq:elbo} sets the logarithm of each factor of $q$ to the expectation of $\log p(\bm Z, \tilde{\bm Z}, \bm W)$ under the other factor, up to an additive constant \citep{bishop2006pattern}.

\paragraph{E-step.} On the half-line prescribed by the label, the terms of \eqref{eq:joint} that contain $\tilde z_{nk}$ give
\begin{equation}
    \log q(\tilde z_{nk}) \;=\; -\tfrac12\,\E_{q(\bw_k)}\!\big[(\tilde z_{nk} - \bw_k^\top\tilde\bx_n)^2\big] + \text{const}
    \;=\; -\tfrac12\,(\tilde z_{nk} - a_{nk})^2 + \text{const},
\end{equation}
because the expectation equals $(\tilde z_{nk} - a_{nk})^2 + \tilde\bx_n^\top\bSigma\tilde\bx_n$ and its second term does not depend on $\tilde z_{nk}$. The factor $q(\tilde z_{nk})$ is therefore a normal distribution with mean $a_{nk}$ and unit variance, truncated to $(0,\infty)$ if $z_{nk} = 1$ and to $(-\infty, 0]$ if $z_{nk} = 0$. The posterior covariance $\bSigma$ does not enter it, which is why the E-step uses the scores at the posterior mean. The first two moments of this truncated normal distribution are
\begin{equation}\label{eq:tn-moments}
    \E_q[\tilde z_{nk}] \;=\; a_{nk} + z_{nk}\,\frac{\phi(a_{nk})}{\Phi(a_{nk})} - (1 - z_{nk})\,\frac{\phi(a_{nk})}{\Phi(-a_{nk})},
    \qquad
    \E_q[\tilde z_{nk}^2] \;=\; 1 + a_{nk}\,\E_q[\tilde z_{nk}].
\end{equation}
The first is \eqref{eq:ez}. The second is required by the parameter expansion of Appendix~\ref{app:px}.

\paragraph{M-step.} The terms of \eqref{eq:joint} that contain $\bw_k$ give
\begin{align}
    \log q(\bw_k) &= -\tfrac12\sum_{n}\E_{q}\!\big[(\tilde z_{nk} - \bw_k^\top\tilde\bx_n)^2\big] - \tfrac12\,(\bw_k - \bmu^0_k)^\top\bLambda_0(\bw_k - \bmu^0_k) + \text{const} \nonumber\\
    &= -\tfrac12\,\bw_k^\top\big(\bLambda_0 + \tilde{\bm X}^\top\tilde{\bm X}\big)\bw_k + \bw_k^\top\big(\bm\eta_{0,k} + \tilde{\bm X}^\top\E_q[\tilde{\bm z}_{:k}]\big) + \text{const},
\end{align}
where $\tilde{\bm z}_{:k}$ is the $k$-th column of $\tilde{\bm Z}$. This is the logarithm of a normal density with the natural parameters \eqref{eq:mstep}. The precision $\bLambda = \bLambda_0 + \tilde{\bm X}^\top\tilde{\bm X}$ depends neither on the class nor on $q(\tilde{\bm Z})$. It is therefore computed once per batch, and only the means change during the inner iterations.

\paragraph{Relation to the Kalman filter.} Substituting $\bLambda_0 = \bLambda - \tilde{\bm X}^\top\tilde{\bm X}$ into $\bmu_k = \bLambda^{-1}\bm\eta_k$ gives the M-step in the form
\begin{equation}\label{eq:kalman-form}
    \bmu_k \;=\; \bmu^0_k + \bSigma\,\tilde{\bm X}^\top\big(\E_q[\tilde{\bm z}_{:k}] - \tilde{\bm X}\bmu^0_k\big).
\end{equation}
This is the update of a Kalman filter with observation matrix $\tilde{\bm X}$, unit observation noise and gain $\bSigma\tilde{\bm X}^\top$, in which the expected auxiliary variables take the place of the observations. In a linear-Gaussian model the observations are given and \eqref{eq:kalman-form} is applied once. Here $\E_q[\tilde{\bm z}_{:k}]$ depends on $\bmu_k$ through \eqref{eq:tn-moments}, so the E-step and \eqref{eq:kalman-form} are iterated.

\paragraph{Value of the bound and its maximizer.} After an E-step, the ELBO \eqref{eq:elbo} evaluates to
\begin{equation}\label{eq:elbo-value}
    \mathcal{L}(q) \;=\; \sum_{n,k}\log\Phi\big((2z_{nk}-1)\,a_{nk}\big) \;-\; \frac{K}{2}\sum_{n}\tilde\bx_n^\top\bSigma\tilde\bx_n \;-\; \sum_{k}\KL\big(q(\bw_k)\,\|\,p(\bw_k)\big).
\end{equation}
The first term is the log-likelihood of the surrogate at the posterior mean, the second is a penalty for the posterior variance of the scores, and the third is the divergence from the prior. With $\bSigma = \bLambda^{-1}$ fixed, \eqref{eq:elbo-value} depends on the means only through
$\sum_{n,k}\log\Phi\big((2z_{nk}-1)\,\bmu_k^\top\tilde\bx_n\big) - \tfrac12\sum_k(\bmu_k - \bmu^0_k)^\top\bLambda_0(\bmu_k - \bmu^0_k)$,
which is strictly concave because $\log\Phi$ is concave. The inner iterations therefore have a single fixed point, and the mean at this fixed point is the maximum-a-posteriori estimate of the $K$ probit regressions under the prior \eqref{eq:prior}. The number of inner iterations $T$ determines how closely this estimate is approached at each batch.

\subsection{Parameter expansion}
\label{app:px}

Parameter expansion adds a scale parameter $\nu$ to the model, which makes the ELBO a function of $\nu$, and maximizes the ELBO over $\nu$ in every inner iteration. The update \eqref{eq:px-update}, the quadratic equation for $\nu$ and its coefficients $c_1$ and $c_2$ all follow from this maximization, as shown below.

\paragraph{Why plain CAVI is slow.} In the exact posterior the overall scale of $\bm W$ and the overall scale of $\tilde{\bm Z}$ are coupled, because $\tilde z_{nk}$ has mean $\bw_k^\top\tilde\bx_n$. The factorization $q(\bm W)\,q(\tilde{\bm Z})$ does not represent this coupling. The E-step computes $q(\tilde{\bm Z})$ for the current scale of $\bm W$ and the M-step computes $q(\bm W)$ for the current scale of $\tilde{\bm Z}$, so the common scale changes by a small amount per iteration. Parameter-expanded variational Bayes \citep{qi2007parameter} introduces a parameter that changes both scales together and optimizes it in every iteration.

\paragraph{Expanded model.} For $\nu > 0$ we define the model $p_\nu$ by
\begin{equation}\label{eq:px-model}
    p_\nu(\bw_k) = \N\big(\bw_k \mid \nu\,\bmu^0_k,\ \nu^2\bSigma_0\big),
    \qquad
    p_\nu(\tilde z_{nk} \mid \bw_k) = \N\big(\tilde z_{nk} \mid \bw_k^\top\tilde\bx_n,\ \nu^2\big),
\end{equation}
with the observation rule $z_{nk} = \mathbf{1}[\tilde z_{nk} > 0]$ unchanged.
It is the original model at $\nu = 1$. For any other value it is the original model in rescaled variables: if $(\bm W, \tilde{\bm Z})$ is distributed according to the original model, then $(\nu\bm W, \nu\tilde{\bm Z})$ is distributed according to $p_\nu$. The labels depend only on the signs of the auxiliary variables, which the rescaling does not change, so $p_\nu(\bm Z) = p(\bm Z)$ for every $\nu$. The parameter $\nu$ therefore does not change the marginal likelihood of the labels.

\paragraph{Expanded ELBO.} The ELBO of the expanded model,
\begin{equation}\label{eq:px-elbo}
    \mathcal{L}(q, \nu) \;=\; \E_q\!\left[\log p(\bm Z \mid \tilde{\bm Z}) + \log p_\nu(\tilde{\bm Z} \mid \bm W) + \log p_\nu(\bm W) - \log q(\bm W, \tilde{\bm Z})\right] \;\le\; \log p(\bm Z),
\end{equation}
is a lower bound on the same marginal likelihood for every $\nu$, and it equals \eqref{eq:elbo} at $\nu = 1$. For a given $q$ it does depend on $\nu$, so it can be increased by maximizing over $\nu$ in addition to the two factors of $q$. Let $q(\tilde{\bm Z})$ be the result of the E-step and let $q(\bw_k) = \N(\bm m_k, \bm S)$. The terms of \eqref{eq:px-elbo} that depend on $\nu$, on the means $\bm m_k$ or on the covariance $\bm S$ are
\begin{equation}\label{eq:px-elbo-terms}
    \mathcal{L} \;=\; -(N+d)K\log\nu \;+\; \frac{K}{2}\log|\bm S| \;-\; \frac{K}{2\nu^2}\,\mathrm{tr}(\bLambda\bm S) \;-\; \frac{1}{2\nu^2}\,R(\nu, \bm m) \;+\; \text{const},
\end{equation}
\begin{equation}\label{eq:px-residual}
    R(\nu, \bm m) \;=\; \sum_{n,k}\E_q\!\big[(\tilde z_{nk} - \bm m_k^\top\tilde\bx_n)^2\big] + \sum_{k}(\bm m_k - \nu\bmu^0_k)^\top\bLambda_0(\bm m_k - \nu\bmu^0_k),
\end{equation}
where $\bLambda = \bLambda_0 + \tilde{\bm X}^\top\tilde{\bm X}$ as in \eqref{eq:mstep} and the expectation in \eqref{eq:px-residual} is over $q(\tilde z_{nk})$ only. The first term of $R$ comes from the likelihood of the auxiliary variables and the second from the prior. We maximize \eqref{eq:px-elbo-terms} over its three arguments in turn.

\paragraph{Covariance.} The maximizer of \eqref{eq:px-elbo-terms} over $\bm S$ is $\bm S = \nu^2\bLambda^{-1}$, whatever the means. With this covariance the trace term is the constant $Kd/2$ and $\tfrac{K}{2}\log|\bm S| = Kd\log\nu + \text{const}$, so that
\begin{equation}\label{eq:px-objective}
    \mathcal{L}(\nu, \bm m) \;=\; -NK\log\nu \;-\; \frac{1}{2\nu^2}\,R(\nu, \bm m) \;+\; \text{const}.
\end{equation}

\paragraph{Means.} For a given $\nu$, $R(\nu, \bm m)$ is quadratic in $\bm m_k$, and \eqref{eq:px-objective} is maximized by
\begin{equation}\label{eq:px-mean}
    \bm m_k(\nu) \;=\; \bLambda^{-1}\big(\nu\,\bm\eta_{0,k} + \bm b_k\big),
    \qquad
    \bm b_k = \tilde{\bm X}^\top\E_q[\tilde{\bm z}_{:k}].
\end{equation}
At $\nu = 1$ this is the plain M-step \eqref{eq:mstep}. We denote its mean by $\hat\bmu_k = \bLambda^{-1}(\bm\eta_{0,k} + \bm b_k)$.

\paragraph{Scale.} Expanding the second term of \eqref{eq:px-residual} in powers of $\nu$ shows that, for given means, \eqref{eq:px-objective} is $-NK\log\nu + c_1/\nu - c_2/(2\nu^2)$ up to a constant, with
\begin{equation}\label{eq:px-coeffs}
    c_1 \;=\; \sum_{k}\bm\eta_{0,k}^\top\bm m_k,
    \qquad
    c_2 \;=\; \sum_{n,k}\E_q\!\big[(\tilde z_{nk} - \bm m_k^\top\tilde\bx_n)^2\big] + \sum_{k}\bm m_k^\top\bLambda_0\bm m_k .
\end{equation}
Its derivative with respect to $\nu$ is $-(NK\nu^2 + c_1\nu - c_2)/\nu^3$. Setting the derivative to zero gives the quadratic equation of \eqref{eq:px-update}, $NK\nu^2 + c_1\nu - c_2 = 0$. Since $c_2 > 0$, the equation has exactly one positive root,
\begin{equation}\label{eq:px-root}
    \nu \;=\; \frac{-c_1 + \sqrt{c_1^2 + 4NK\,c_2}}{2NK},
\end{equation}
and this root is the maximizer, because the derivative is positive below it and negative above it.

\paragraph{One inner iteration.} Each inner iteration of Algorithm~\ref{alg:cavi} starts from the original model, $\nu = 1$, with the current mean $\bmu$, and performs the following steps.
\begin{enumerate}
    \item \emph{E-step.} Compute the moments \eqref{eq:tn-moments} of $q(\tilde{\bm Z})$ at the scores $a_{nk} = \bmu_k^\top\tilde\bx_n$, and the data term $\bm b_k$.
    \item \emph{Means at $\nu = 1$.} Set $\bm m_k = \hat\bmu_k$, the mean of the plain M-step.
    \item \emph{Scale.} Set $\nu$ to the root \eqref{eq:px-root}, with $c_1$ and $c_2$ evaluated at $\bm m_k = \hat\bmu_k$.
    \item \emph{Means at the new $\nu$.} Set $\bm m_k = \bLambda^{-1}(\nu\,\bm\eta_{0,k} + \bm b_k)$ by \eqref{eq:px-mean}. The covariance is $\nu^2\bLambda^{-1}$.
    \item \emph{Reduction.} Return to the original model by the change of variables $(\bm W, \tilde{\bm Z}) \to (\bm W/\nu,\ \tilde{\bm Z}/\nu)$, which maps $p_\nu$ to $p_1$ and divides the means by $\nu$ and the covariance by $\nu^2$:
    \begin{equation}\label{eq:px-reduced}
        q(\bw_k) \;=\; \N\big(\bLambda^{-1}(\bm\eta_{0,k} + \bm b_k/\nu),\ \bLambda^{-1}\big).
    \end{equation}
\end{enumerate}
The result \eqref{eq:px-reduced} is the update \eqref{eq:px-update}: the first natural parameter is $\bm\eta_k = \bm\eta_{0,k} + \bm b_k/\nu$ and the precision $\bLambda$ is unchanged. In comparison with the plain M-step, the only difference is that the data term $\bm b_k$ is divided by $\nu$. Step 1 is line 8 of Algorithm~\ref{alg:cavi}, and steps 2 to 5 together are line 9.

\paragraph{The bound does not decrease.} Steps 2 to 4 each maximize \eqref{eq:px-objective} over one of its arguments, so none of them decreases the bound. Step 5 leaves the value of the bound unchanged, because a change of variables applied to both the model and $q$ does not change the divergence between them. The reduction also rescales $q(\tilde{\bm Z})$, which the E-step of the next iteration replaces by the optimal factor, again without decreasing the bound. The ELBO \eqref{eq:elbo} of the original model therefore does not decrease from one inner iteration to the next, and after each iteration it is at least as large as it would be after the plain M-step from the same $q(\tilde{\bm Z})$.

\paragraph{Fixed point.} At a fixed point of plain CAVI the M-step returns the mean at which the E-step was evaluated, $\hat\bmu_k = \bmu_k$. By \eqref{eq:tn-moments}, $\E_q[(\tilde z_{nk} - a_{nk})^2] = 1 - a_{nk}\big(\E_q[\tilde z_{nk}] - a_{nk}\big)$, and summing over $n$ and $k$ gives $NK - \sum_k\bmu_k^\top\big(\bm b_k - \tilde{\bm X}^\top\tilde{\bm X}\bmu_k\big)$. The M-step \eqref{eq:mstep} gives $\bm b_k - \tilde{\bm X}^\top\tilde{\bm X}\bmu_k = \bLambda_0\bmu_k - \bm\eta_{0,k}$. Substituting both into \eqref{eq:px-coeffs} gives
\begin{equation}
    c_2 - c_1 \;=\; NK,
\end{equation}
so $\nu = 1$ solves the quadratic equation and the iteration leaves the fixed point unchanged. Conversely, at a fixed point of the PX iteration none of the steps increases the bound, so step 2 leaves the means unchanged and the point is a fixed point of plain CAVI. The two algorithms therefore have the same fixed point, which by Appendix~\ref{app:cavi} is unique, and they differ only in the number of iterations needed to approach it. Away from the fixed point $\nu \neq 1$ in general.

\paragraph{Form used in the implementation.} With $\hat{\bm\eta}_k = \bm\eta_{0,k} + \bm b_k$ and the scores $\hat a_{nk} = \hat\bmu_k^\top\tilde\bx_n$ of the plain M-step, the coefficients \eqref{eq:px-coeffs} at $\bm m_k = \hat\bmu_k$ are
\begin{align}\label{eq:px-coeffs-impl}
    c_1 &= \sum_{k}\big\langle\bLambda^{-1}\bm\eta_{0,k},\ \hat{\bm\eta}_k\big\rangle, \nonumber\\
    c_2 &= \sum_{n,k}\Big(\E_q[\tilde z_{nk}^2] - 2\,\E_q[\tilde z_{nk}]\,\hat a_{nk} + \hat a_{nk}^2\Big) + \sum_{k}\big\langle\hat{\bm\eta}_k,\ \bLambda^{-1}\bLambda_0\bLambda^{-1}\hat{\bm\eta}_k\big\rangle,
\end{align}
with $\E_q[\tilde z_{nk}^2]$ from \eqref{eq:tn-moments}. The matrices $\bLambda^{-1}\bm\eta_0$ and $\bLambda^{-1}\bLambda_0\bLambda^{-1}$ do not change during the inner iterations. They are computed once per batch from the Cholesky factorization of $\bLambda$ that the M-step already requires, so each inner iteration consists of matrix multiplications only.

\subsection{The filtering recursion and its steady state}
\label{app:filter}

\paragraph{Prediction step.} Let $\bw_{m-1,k} \sim \N(\bmu_{m-1,k}, \bSigma_{m-1})$ and $\bw_{m,k} = \bw_{m-1,k} + \bm\xi_{m,k}$, where $\bm\xi_{m,k} \sim \N(\bm 0, \bQ)$ is independent of $\bw_{m-1,k}$. Then $\bw_{m,k} \sim \N(\bmu_{m-1,k}, \bSigma_{m-1} + \bQ)$, which is \eqref{eq:predict}. This step involves no approximation. Because $\bQ$ is the same for every class, the predicted covariance is again shared across classes, and the predicted distribution has the form of the prior \eqref{eq:prior} that the update of Appendix~\ref{app:cavi} requires.

\paragraph{Covariance recursion.} Combining \eqref{eq:predict} with the precision of \eqref{eq:mstep} gives
\begin{equation}\label{eq:cov-recursion}
    \bSigma_m \;=\; \Big[\big(\bSigma_{m-1} + q\bI\big)^{-1} + \tilde{\bm X}_m^\top\tilde{\bm X}_m\Big]^{-1}.
\end{equation}
The covariance depends on the features only, and not on the labels or on the inner iterations. To analyse the recursion we consider the case in which every batch has the same Gram matrix, $\tilde{\bm X}_m^\top\tilde{\bm X}_m = \bm G$. Starting from $\bSigma_0 = \bI$, the matrix $\bSigma_m$ then has the eigenvectors of $\bm G$ for every $m$, and each of its eigenvalues follows the scalar recursion
\begin{equation}\label{eq:scalar-recursion}
    \Sigma_m \;=\; \big[(\Sigma_{m-1} + q)^{-1} + \lambda\big]^{-1},
\end{equation}
where $\lambda$ is the corresponding eigenvalue of $\bm G$, that is, the precision that one batch contributes in that direction. We refer to \eqref{eq:scalar-recursion} as the scalar reduction. For batches that differ from one another it describes the average behaviour of the filter, with $\lambda$ the average precision per batch, and the statements below hold in this approximate sense.

\paragraph{Without process noise.} For $q = 0$ and $\Sigma_0 = 1$, \eqref{eq:scalar-recursion} gives $\Sigma_m = (1 + m\lambda)^{-1}$, which decreases to zero as the number of batches increases.

\paragraph{Lower bound on the covariance.} For $q > 0$ the recursion converges to the positive solution of $\lambda\Sigma^{*2} + \lambda q\,\Sigma^{*} - q = 0$,
\begin{equation}\label{eq:sigma-fp}
    \Sigma^{*} \;=\; \frac{q}{2}\left(\sqrt{1 + \frac{4}{\lambda q}} - 1\right)
    \;\approx\;
    \begin{cases}
        \sqrt{q/\lambda}, & \lambda q \ll 1,\\[2pt]
        1/\lambda, & \lambda q \gg 1.
    \end{cases}
\end{equation}
In the first regime the limit is the geometric mean of the process noise $q$ and the variance $1/\lambda$ of the estimate from a single batch, and not $q$ itself. In the second regime the process noise is so large that the previous posterior contributes no information in this direction, and the posterior is that of the current batch alone.

\paragraph{Gain and memory.} In the scalar reduction, the update \eqref{eq:kalman-form} with the prior mean $\mu_{m-1}$ of the prediction step reads
\begin{equation}\label{eq:scalar-mean}
    \mu_m \;=\; (1 - g_m)\,\mu_{m-1} + g_m\,\frac{b_m}{\lambda},
    \qquad
    g_m = \lambda\,\Sigma_m,
\end{equation}
where $b_m$ is the data term of batch $m$ and $b_m/\lambda$ is the estimate that this batch alone would give. At the steady state the weight of the newest batch is $g^{*} = \lambda\Sigma^{*} \approx \sqrt{\lambda q}$ for $\lambda q \ll 1$, and the data term of a batch that was assimilated $j$ steps earlier enters the mean with the weight $g^{*}(1 - g^{*})^{j}$. The number of filtering steps over which a batch influences the posterior mean is therefore
\begin{equation}\label{eq:memory}
    \tau \;=\; \frac{1}{g^{*}} \;\approx\; \frac{1}{\sqrt{\lambda q}}.
\end{equation}
The process noise thus determines both the lower bound \eqref{eq:sigma-fp} on the posterior covariance and the memory \eqref{eq:memory} of the filter.

\paragraph{The effective learning rate.} The gain $g_m$ in \eqref{eq:scalar-mean} is the learning rate of the mean: each batch moves the mean by the fraction $g_m$ of its distance to the estimate of that batch. The precision that a batch of $B$ inputs contributes is $\lambda = Bc$, where $c$ is the second moment of the features in the direction considered. With the parametrization $q = \alpha^2/B$ of \eqref{eq:q-prior},
\begin{equation}\label{eq:alpha-gain}
    g^{*} \;\approx\; \sqrt{\lambda q} \;=\; \alpha\sqrt{c},
    \qquad
    \tau \;\approx\; \frac{1}{\alpha\sqrt{c}}.
\end{equation}
For features of unit scale per coordinate $c = 1$ in every direction, so $\alpha$ is the steady-state gain itself and $1/\alpha$ is the memory of the filter in steps, which is why we call $\alpha$ the effective learning rate of the head. For the features used in the experiments, which are not normalised (Appendix~\ref{sec:appendix-setup}), $c$ differs between directions, and $\alpha$ is the gain up to the factor $\sqrt{c}$. Appendix~\ref{sec:appendix-q} uses \eqref{eq:memory} to derive how $q$ has to change with the number of training steps.

\subsection{The posterior predictive}
\label{app:moderated}

\paragraph{Expectation of the probit function.} Let $u \sim \N(\bar u, \sigma_u^2)$ and let $\epsilon \sim \N(0, 1)$ be independent of $u$. Then $\Phi(u) = \Pr(\epsilon \le u \mid u)$, and
\begin{equation}\label{eq:probit-gauss}
    \E\big[\Phi(u)\big] \;=\; \Pr(\epsilon - u \le 0) \;=\; \Phi\!\left(\frac{\bar u}{\sqrt{1 + \sigma_u^2}}\right),
\end{equation}
because $\epsilon - u \sim \N(-\bar u,\, 1 + \sigma_u^2)$. This identity is exact.

\paragraph{Moderated scores.} A new input $\tilde\bx$ is scored by the head one transition after its last update, so the weights have the predicted distribution $\bw_k \sim \N(\bmu_k, \bSigma + \bQ)$ of \eqref{eq:predict}. The score $u = \bw_k^\top\tilde\bx$ is then normally distributed with mean $\bmu_k^\top\tilde\bx$ and variance $\tilde\bx^\top(\bSigma + \bQ)\tilde\bx = \tilde\bx^\top\bSigma\tilde\bx + q\|\tilde\bx\|^2$, and \eqref{eq:probit-gauss} gives \eqref{eq:moderated}. Neither $\bSigma$ nor $\bQ$ depends on the class, so $s(\tilde\bx)$ is one positive number per input. Dividing all $K$ scores by it does not change their order, and hence does not change the predicted class. It changes only the confidence of the prediction.

\paragraph{Class probabilities.} The factor $q(\bm W)$ and the transition \eqref{eq:transition} are both independent across classes. Under the predictive distribution of the surrogate, the $K$ binary variables of an input are therefore independent, with probabilities $\Phi(a_k(\tilde\bx))$, where $a_k(\tilde\bx) = \bmu_k^\top\tilde\bx/s(\tilde\bx)$. Conditioning this distribution on exactly one active class, as in Appendix~\ref{app:cb-bound}, gives the class probabilities \eqref{eq:cbc-loss}. They are the predictive distribution of the surrogate conditioned on one active class. They are not the expectation of \eqref{eq:cbc} under $q(\bm W)$, which has no closed form.

\subsection{Computational cost of a filtering step}
\label{app:cost}

One filtering step on a batch of $B$ inputs consists of the following operations. The prediction step factorizes $\bSigma_{m-1} + \bQ$ and solves for $\bLambda^{\mathrm{pred}}_m$ and $\bm\eta^{\mathrm{pred}}_m$, at a cost of $\mathcal{O}(d^3 + d^2K)$. The update forms $\bLambda_m = \bLambda^{\mathrm{pred}}_m + \tilde{\bm X}_m^\top\tilde{\bm X}_m$, factorizes it, and computes $\tilde{\bm X}_m\bLambda_m^{-1}$ and the two constant matrices of \eqref{eq:px-coeffs-impl}, at a cost of $\mathcal{O}(Bd^2 + d^3 + d^2K)$. Each of the $T$ inner iterations then computes the scores $\tilde{\bm X}_m\bLambda_m^{-1}\bm\eta$, the data term and the coefficients $c_1$ and $c_2$ by matrix multiplications, at a cost of $\mathcal{O}(BdK + d^2K)$, of which the second term is due to the parameter expansion. Finally $\bmu_m$ and $\bSigma_m$ are obtained from the factorization of $\bLambda_m$. The total cost of a step is
\begin{equation}
    \mathcal{O}\big(d^3 + Bd^2 + T\,(BdK + d^2K)\big),
\end{equation}
with two Cholesky factorizations per batch, whose cost depends neither on $T$ nor on $K$, and an inner loop that is linear in $K$.

The state of the head consists of the means, with $Kd$ entries as in any linear head, and the shared covariance, with $d(d+1)/2$ distinct entries for any number of classes. A per-class diagonal covariance, as in the diagonal parameterization of VBLL, has $Kd$ entries, and a per-class factorization of rank $r$ has $Kdr$.

\section{Experimental setup}
\label{sec:appendix-setup}

This appendix describes what is common to all experiments. The protocols of the probing and detection experiments and of the fine-tuning experiment are given in Appendices~\ref{sec:appendix-tuning} and~\ref{sec:appendix-tuning-ft}.

\paragraph{Backbones and features.} All experiments use DINOv3 \citep{simeoni2025dinov3}, in the four sizes ViT-S+/16, ViT-B/16, ViT-L/16 and ViT-H+/16, which we refer to as Small, Base, Large and Huge and which produce features of dimension $D = 384$, $768$, $1024$ and $1280$. The feature $\bx$ is the mean of the patch tokens of the final layer. The features are used as the backbone produces them, without normalisation; for DINOv3-Huge their mean square per coordinate is between $0.01$ and $0.06$ on the probing datasets.

\paragraph{Datasets.} The probing experiments (\S\ref{sec:probing-scale}) use Oxford Pets, Food-101, Flowers-102, Stanford Cars and DTD, and the detection experiments (\S\ref{sec:ood-detection}) use CIFAR-10, CIFAR-100 and SVHN. For these experiments the features are extracted once per backbone and dataset, from images resized to $512 \times 512$ pixels for the five probing datasets and to $224 \times 224$ pixels for CIFAR-10/100 and SVHN, and stored. Every method therefore receives identical inputs, and none can benefit from different preprocessing. The fine-tuning experiment (\S\ref{sec:fullnetwork}) trains ViT-H+/16 end to end on ImageNet-1k \citep{deng2009imagenet} at $224 \times 224$ pixels, without data augmentation, and evaluates under shift on ImageNet-R, -A, -V2 and -Sketch.

\paragraph{Methods.} IBProbit is the head of \S\ref{sec:cavi}--\S\ref{sec:loss-formulations}. Every run uses the parameter expansion of \S\ref{sec:cavi}, the loss \eqref{eq:cbc-loss} with the approximation \eqref{eq:approx-cdf}, the initial state $\bmu = \bm 0$, $\bSigma = \bI$, and a process noise $\bQ = q\bI$ that is constant during a run. Its hyperparameters are the process noise $q$, the number of inner iterations $T$ and the batch size. AdamW+CE is a linear layer trained with the cross-entropy loss, and CE+TS applies to it a single temperature that is fitted on the validation set after training. VBLL is the classification head of \citet{harrison2024variational} in its diagonal parameterization. Its predictive is the Monte Carlo estimate of the expected softmax, with $200$ samples in the probing experiments and $20$ in fine-tuning.

\paragraph{Optimization and implementation.} The gradient-trained heads (AdamW+CE and VBLL) and, in fine-tuning, the backbone are trained with AdamW \citep{kingma2014adam, loshchilov2019decoupled} at a constant learning rate. The cross-entropy and IBProbit experiments use JAX \citep{jax2018github} with Equinox \citep{kidger2021equinox}; the VBLL experiments use the PyTorch implementation of \citet{harrison2024variational}. In probing all methods read the same stored features; in fine-tuning the VBLL runs use a PyTorch implementation of the same pretrained backbone with the same pooled feature.

\paragraph{Metrics and seeds.} We report the accuracy, the negative log-likelihood (NLL) and the multiclass Brier score, which is computed over the full predictive vector. Detection is scored by the maximum softmax probability, for which we report the AUROC and the false-positive rate at a true-positive rate of $95\%$. Test-set numbers are averaged over seeds: ten for the linear-probing and detection experiments (\S\ref{sec:probing-scale}, \S\ref{sec:ood-detection}) and five for full-network fine-tuning (\S\ref{sec:fullnetwork}). Figures~\ref{fig:probing-scale} and~\ref{fig:ood-detection} show the seed mean with error bars of $\pm 1$ standard deviation; in Figures~\ref{fig:sample-efficiency} and~\ref{fig:shift} the bands are the range across seeds.

\paragraph{Compute resources.} The reported linear-probing and detection
runs (\S\ref{sec:probing-scale}, \S\ref{sec:ood-detection}: 690 runs across three methods, four backbone sizes, seven datasets and ten seeds) total approximately 250
GPU-hours on NVIDIA A100s. The reported full-network fine-tuning runs
(\S\ref{sec:fullnetwork}: three methods $\times$ five seeds $\times$ two epochs of
ImageNet-1k on DINOv3-Huge) total approximately 280 GPU-hours on the same
hardware, $18$--$19$ hours per run, plus roughly $40$ GPU-hours for the separate
workers that score the checkpoints of the PyTorch runs on the shift sets. The
hyperparameter searches behind both cost substantially more than the reported
runs: 21 studies of roughly 100 completed trials each for probing, and
approximately $930$ GPU-hours for the three DINOv3-Huge fine-tuning studies whose
configurations are reported (Appendix~\ref{sec:ft-search}). The project
as a whole cost considerably more again, through the DINOv3-Large studies, the
process-noise measurements of Appendix~\ref{sec:appendix-q},
preliminary exploration, abandoned configurations and earlier algorithm
variants.

\section{Linear probing and detection: protocol, tuning and final runs}
\label{sec:appendix-tuning}

Figures~\ref{fig:probing-scale} and~\ref{fig:ood-detection} compare three heads,
so the comparison is only as reliable as the tuning of the two baselines. This appendix describes the protocol of these experiments in the order in which it is carried out: the training procedure, the data splits and the budget, then the hyperparameter search, the final runs behind every plotted point, and the caveats that apply to the figures. One protocol is applied identically to every method. The fine-tuning experiment of \S\ref{sec:fullnetwork} was tuned separately and is described in Appendix~\ref{sec:appendix-tuning-ft}.

\paragraph{Training procedure.} Each method is trained on the stored features of the training split of a dataset. AdamW+CE and VBLL take one AdamW step per mini-batch. IBProbit takes one filtering step per mini-batch (Algorithm~\ref{alg:cavi}), which requires no gradient computation. The filter makes repeated passes over the training data, and its posterior is carried from one pass to the next without a reset.

\paragraph{Data splits.} Training uses the full training split. A fixed $10\%$ of the held-out split of each dataset, the same for every method and seed, is the validation set: it is used to select the evaluation at which a run is scored, to score the search trials and to fit the temperature. All reported numbers are computed on the remaining $90\%$. For Flowers-102 the training split is the official one of $1{,}020$ images and the held-out split is the union of the official validation and test sets.

\paragraph{Budget and evaluation.} Every run of every method has the same budget of $30{,}000$ steps, expressed as whole epochs at that run's batch size, so two configurations with different batch sizes receive the same number of updates rather than the same number of passes. The validation NLL is evaluated on a grid that is uniform in the number of training \emph{samples}, with the same interval for every batch size of a dataset. A run is scored at the evaluation with the lowest validation NLL.

\paragraph{Search spaces.} Each method is searched over its own hyperparameters. AdamW+CE: learning rate $[10^{-3}, 3\times10^{-2}]$ and weight
decay $[10^{-5}, 10^{-1}]$, both log-uniform, plus batch size over a per-dataset set of powers of two. VBLL adds its three variational hyperparameters --- KL ratio
$[10^{-7},10^{-2}]$, prior scale $[10^{-1},10^{3}]$, Wishart scale
$[10^{-3},10^{1}]$, all log-uniform --- for six searched dimensions, the largest
of the three. IBProbit has neither a learning rate nor a weight decay, because the
head is updated in closed form and the backbone is frozen; it is searched over
$q$ (log-uniform in $[10^{-6},10^{-1}]$), the number of inner iterations
$T \in \{32,64,128,192\}$, and batch size. Temperature scaling is not searched ---
it is fit post hoc on the validation split, so CE+TS uses none of the CE search budget and improves the baseline at no additional search cost.

\paragraph{Objective, and the smoothing window.} A trial scores the lowest mean
validation NLL over any two consecutive evaluations, rather than the single lowest
value. Each trial is run with two seeds, neither of which is among the ten reported ones, and its score is the mean over the two. The width of the window affects which configurations are selected. On a trajectory that does not converge, the single-point minimum reflects the amplitude of the noise rather than the level reached, and the search favours configurations whose validation loss oscillates strongly enough to reach low values: on Oxford Pets the unsmoothed objective selected a CE learning
rate of $0.0182$ whose late-phase standard deviation is $0.53\times$ its own
median, over a configuration $13\times$ more stable and $0.032$ better on test
NLL.

Only methods with a learning rate can benefit from this, so the bias is not symmetric across the methods being compared. A window that is too wide has a different undesired effect, on IBProbit. We measured both by repeating the entire search at each width. Widths $1$ and $2$ select the same operating point for IBProbit
(within $0.003$ NLL at every task and $0.005$ FPR@95 at every pair), while width
$5$ selects $q$ at its lower search bound, $1.0\times10^{-6}$ --- confirmed on an
independent search seed --- and increases FPR@95 by a factor of three on CIFAR-10 vs SVHN. Conversely the gradient-trained methods need width $\geq 2$ to avoid the selection of noisy minima described above. Width $2$ is the only setting satisfying both, and it is
applied uniformly to all three methods; no method is tuned under an objective of its own.

\paragraph{Search procedure and depth.} Each (method, dataset) study is a separate
Optuna \citep{akiba2019optuna} TPE search run by four parallel workers against a
shared journal, with a median pruner (20 startup trials, 10 warmup evaluations). In addition, a trial is
stopped early after 40 consecutive evaluations without relative improvement.
Workers stop once the study holds 100 completed trials, counting across workers,
so the search depth is fixed in advance and does not depend on how many trials the pruner stopped. Of the 21 studies, 17 reached that
target. Four stopped short --- IBProbit on Oxford Pets (73 completed) and Stanford
Cars (86), CE (79) and VBLL (89) on CIFAR-10 --- because prune rates in those studies reach $83$--$93\%$ and the per-worker trial limit was reached first. We report the shortfall instead of adding trials: in the two CIFAR-10 studies a further 600 trials yielded only nine additional completions, which indicates a converged study rather than one with too few trials. Table~\ref{tab:tuned-configs} lists every selected configuration.

\paragraph{Final runs.} Reported numbers come from new runs at the selected
configuration, never from the search trials. Each final run uses the full $30{,}000$-step budget, without the early stopping of the search, and is scored at the evaluation with the lowest validation NLL. At DINOv3-Huge this evaluation falls, for IBProbit, after a median of $13$ (DTD) to $686$ (Flowers-102) passes over the training data. Figure~\ref{fig:probing-scale} is 3 methods $\times$ 4 backbone sizes $\times$ 5 tasks $\times$ 10 seeds $= 600$ runs, all of which completed. The figure uses 598 of them: one seed each of CE on Food-101 and of
IBProbit on Flowers-102 at DINOv3-Base was excluded by the analysis script, which carried over an exclusion list from an earlier campaign in which these two runs had failed, so those two settings have nine seeds.
Figure~\ref{fig:ood-detection} is 3 methods $\times$ 3 pairs $\times$ 10 seeds $= 90$
runs, all of which completed.

\paragraph{Caveats.} There are three, all of which affect how the figures should be interpreted.
\emph{Only DINOv3-Huge was tuned.} The other three backbone sizes use the configuration tuned at Huge,
and the backbones produce 384/768/1024/1280-dimensional features, so a configuration
optimal at one is not guaranteed optimal at another; the non-Huge points in
Figure~\ref{fig:probing-scale} are therefore not separately optimized for any method,
which is a limitation applying equally to all three. \emph{69 of the 598 runs are
budget-truncated}, meaning their validation optimum falls in the last 10\% of the
step budget and the plotted point is a lower bound; these concentrate at
DINOv3-Small and on Food-101, and the figure marks them. \emph{The IBProbit configuration on Oxford Pets is close to a region where training diverges}: 30 of the 600 trials in that study
produced a non-finite validation NLL at every evaluation, all at batch 1024 with
$T=192$, which is also the setting of the selected configuration. The
selected configuration itself completed in all ten reported seeds, but roughly one trial in nine with these settings fails.

\begin{table}[ht!]
\centering
\caption{Selected configurations, one per (method, dataset), tuned at DINOv3-Huge and applied at every backbone size. \texttt{epochs} restates the shared 30{,}000-step
budget at that entry's batch size and is not an independent hyperparameter. The final column
is the number of completed trials the study reached. IBProbit has no learning rate
or weight decay: its head update is closed-form.}
\label{tab:tuned-configs}
\small
\begin{tabular}{lccccc}
\toprule
\multicolumn{6}{l}{\textit{AdamW + cross-entropy} (+ post-hoc temperature scaling)} \\
Dataset & batch & learning rate & weight decay & epochs & trials \\
\midrule
Oxford Pets & 64 & $0.00395$ & $2.64\times 10^{-5}$ & 527 & 102 \\
Food-101 & 512 & $0.00113$ & $0.000223$ & 205 & 100 \\
Flowers-102 & 256 & $0.021$ & $0.00168$ & 10000 & 103 \\
Stanford Cars & 64 & $0.00101$ & $1.6\times 10^{-5}$ & 237 & 102 \\
DTD & 64 & $0.00217$ & $0.00103$ & 518 & 100 \\
CIFAR-10 & 512 & $0.0146$ & $0.072$ & 310 & 79 \\
CIFAR-100 & 512 & $0.00167$ & $0.000132$ & 310 & 101 \\
\midrule
\multicolumn{6}{l}{\textit{IBProbit}} \\
Dataset & batch & $q$ & $T$ & epochs & trials \\
\midrule
Oxford Pets & 1024 & $0.00139$ & 192 & 10000 & 73 \\
Food-101 & 1024 & $1.93\times 10^{-6}$ & 64 & 411 & 103 \\
Flowers-102 & 256 & $9.52\times 10^{-5}$ & 32 & 10000 & 103 \\
Stanford Cars & 256 & $2.58\times 10^{-5}$ & 192 & 968 & 86 \\
DTD & 64 & $0.00101$ & 192 & 518 & 102 \\
CIFAR-10 & 64 & $0.00577$ & 128 & 39 & 102 \\
CIFAR-100 & 128 & $0.000529$ & 32 & 77 & 102 \\
\bottomrule
\end{tabular}

\vspace{1em}

\begin{tabular}{lccccccc}
\toprule
\multicolumn{8}{l}{\textit{VBLL} (diagonal parameterization)} \\
Dataset & batch & lr & weight decay & KL ratio & prior scale & Wishart scale & trials \\
\midrule
Oxford Pets & 128 & $0.0159$ & $7.61\times 10^{-5}$ & $2.36\times 10^{-7}$ & $1.45$ & $0.546$ & 101 \\
Food-101 & 128 & $0.00127$ & $0.000566$ & $1.15\times 10^{-6}$ & $0.103$ & $0.00555$ & 100 \\
Flowers-102 & 64 & $0.0295$ & $0.00597$ & $3.81\times 10^{-7}$ & $11.1$ & $0.371$ & 102 \\
Stanford Cars & 64 & $0.0034$ & $0.000126$ & $1.42\times 10^{-7}$ & $12.6$ & $0.0201$ & 100 \\
DTD & 64 & $0.0072$ & $0.0377$ & $1.67\times 10^{-5}$ & $16.5$ & $0.0463$ & 103 \\
CIFAR-10 & 1024 & $0.0142$ & $0.00254$ & $0.00396$ & $0.124$ & $0.371$ & 89 \\
CIFAR-100 & 512 & $0.00179$ & $0.0775$ & $9.85\times 10^{-6}$ & $6.16$ & $6.94$ & 102 \\
\bottomrule
\end{tabular}
\end{table}

\section{Full-network fine-tuning: procedure, tuning and transfer}
\label{sec:appendix-tuning-ft}

This appendix describes the fine-tuning experiment of \S\ref{sec:fullnetwork} in the order in which it is carried out: the training step (Appendix~\ref{sec:ft-procedure}), the hyperparameter search at a short training duration (Appendix~\ref{sec:ft-search}), the transfer of the selected hyperparameters to the reported duration, which includes the scaling law for the process noise and its empirical validation (Appendix~\ref{sec:appendix-q}), and the final runs (Appendix~\ref{sec:ft-final}). We write $M$ for the number of training steps of a run. For IBProbit every training step contains one filtering step, so $M$ is also the number of filtering steps.

\subsection{Training procedure}
\label{sec:ft-procedure}

\paragraph{Common settings.} All three methods fine-tune all weights of DINOv3-Huge on ImageNet-1k with AdamW at a constant learning rate. The batch size is $512$, reached by accumulating the gradients of micro-batches of $16$ images ($32$ for VBLL). Layer-wise learning-rate decay reduces the learning rate by a constant factor from each of the $32$ blocks of the backbone to the block below it, such that the learning rate of the input embedding is $0.75^{12}$ times that of the last block. The backbone is evaluated in bfloat16 precision, and the weights, the optimizer state and the head are kept in float32.

\paragraph{The training step of IBProbit.} One training step on a batch $\mathcal{B}_m$ consists of three operations.
\begin{enumerate}
    \item \emph{Loss and gradient.} The backbone computes the features of the batch, and the loss \eqref{eq:cbc-loss} is evaluated with the head of the previous step, $(\bmu_{m-1}, \bSigma_{m-1})$. The gradient of the loss is taken with respect to the weights of the backbone only. It passes through the features into both the class scores $\bmu_k^\top\tilde\bx$ and the factor $s(\tilde\bx)$ of \eqref{eq:moderated}, and the head is a constant in this computation.
    \item \emph{Backbone step.} AdamW updates the backbone with the gradient averaged over the batch.
    \item \emph{Head step.} The features computed in the first operation, together with the labels, are assimilated by one filtering step of Algorithm~\ref{alg:cavi} on the whole batch of $512$, which gives $(\bmu_m, \bSigma_m)$.
\end{enumerate}
The loss of a batch is thus evaluated before the head has assimilated that batch. In the first step the head is in its initial state $\bmu = \bm 0$, for which the loss does not depend on the features, so only the head step is performed. The head is not reset between epochs: the process noise discounts the evidence obtained from features that earlier states of the backbone produced.

\paragraph{The training step of the baselines.} For AdamW+CE and VBLL the head is trained jointly with the backbone by the same AdamW steps. The learning rate of the head is a multiple of that of the backbone, and this multiple is a searched hyperparameter.

\paragraph{Data splits.} A fixed partition of $10{,}000$ images, $20\%$ of the official ImageNet validation set, is the validation set of this experiment: it is used to score the search trials and to fit the temperature of CE+TS. The remaining $40{,}000$ images are the test set on which every figure and table reports.

\subsection{Hyperparameter search}
\label{sec:ft-search}

\paragraph{Search setting.} Each method is tuned at the reported backbone,
DINOv3-Huge, at a short training duration: $M_{\mathrm{s}} = 781$ steps
($399{,}872$ samples, about a third of an epoch), with the validation NLL evaluated every $2{,}048$
samples. Each study is a TPE search with a median pruner, and each trial uses one seed, which is not among the reported ones.

\paragraph{Search spaces.} The searched coordinates are chosen to be invariant under the
transfer to the reported duration (Appendix~\ref{sec:appendix-q}), so that the selected values remain valid after the transfer. Learning rate and weight decay are searched per
$\sqrt{\text{batch}}$, and the AdamW moment decays as the products $(1-\beta_i)\,M$, which we call horizons, with the $\beta_1$-horizon in $[39, 312]$ and the $\beta_2$-horizon in $[0.4, 40]$. At
batch $512$ the learning-rate search range is $[10^{-6}, 10^{-3}]$ for every method, and the weight-decay search range is $[10^{-6}, 10^{-3}]$ for cross-entropy, $[10^{-5}, 10^{-2}]$
for VBLL and $[10^{-5}, 10^{-1}]$ for IBProbit. Cross-entropy and VBLL also search
the head's learning rate as a multiple of the backbone's, $[5, 10^4]$; VBLL adds its
KL ratio $[10^{-7}, 10^{-2}]$ and prior scale $[10^{-3}, 10^{3}]$. IBProbit has no
head learning rate. It searches $q \in [10^{-7}, 10^{-2}]$ with the number of inner
iterations fixed at $T=256$, since at a value of $q$ appropriate for the training duration
$128$ and $256$ inner iterations gave validation NLL within $0.001$ of each other
(DINOv3-Large, $781$ steps, one seed).

\paragraph{Objective.} Every reported configuration is the trial with the lowest
mean validation NLL over its last five evaluations. A trial is also stopped when its validation NLL has not improved by a relative $10^{-4}$ over $20$ consecutive evaluations; such a trial counts as completed and is scored on its last five evaluations. Scoring the end of the trajectory rather than its minimum is important for a method with a forgetting rate: a search that scores the minimum of a truncated curve prefers configurations whose validation loss is low early and then degrades, which selects a process noise that is too large
(Appendix~\ref{sec:appendix-q}). The IBProbit study was run under this objective
directly, with pruning deferred to $30\%$ of the run, and completed $52$ trials.
The cross-entropy and VBLL studies were run earlier with pruning after ten
evaluations and completed $61$ and $59$ trials; their reported configurations are
selected by applying the same objective to those completed trials. Both searches had converged --- among their near-best trials every cross-entropy
coordinate lies within a factor $2.1$ and every VBLL coordinate within $1.4$ ---
so rerunning them at IBProbit's smaller completion target would have given the
baselines a weaker search, not a fairer one.

\subsection{Transfer to the reported training duration}
\label{sec:appendix-q}

The search is run at $M_{\mathrm{s}} = 781$ steps and the reported runs at $M_{\mathrm{r}} = 5{,}004$ steps, which is two epochs, at the same batch size. Hyperparameters whose appropriate value depends on the number of steps have to be converted from one duration to the other. We first state the two rules that we use, and then report the measurements that test the second of them.

\paragraph{AdamW hyperparameters.} For the hyperparameters of AdamW we use the scaling rule derived from its stochastic-differential-equation limit \citep{mlodozeniec2025completed}. The learning rate and the weight decay are multiplied by $\sqrt{M_{\mathrm{s}}/M_{\mathrm{r}}}$. The moment decays are set such that the horizons $(1-\beta_i)\,M$ are unchanged, which keeps the averaging window $1/(1-\beta_i)$ of each moment estimate at a fixed fraction of the run. The ratio of the head and backbone learning rates is unchanged.

\paragraph{Process noise: the scaling law.} The rule for AdamW does not cover the process noise $q$, whose appropriate value also depends on the number of steps. By \eqref{eq:memory}, a batch influences the posterior mean of the head over $\tau \approx (\lambda q)^{-1/2}$ filtering steps, where $\lambda$ is the precision that one batch contributes. We require that $\tau$ is the same fraction of the run at both durations. This is the requirement that the rule above places on the averaging windows of AdamW, applied to the head. At a fixed batch size $\lambda$ does not change, so $\tau/M$ is unchanged if and only if
\begin{equation}\label{eq:q-scaling}
    q \;\propto\; M^{-2},
    \qquad\text{that is,}\qquad
    q_{\mathrm{r}} \;=\; q_{\mathrm{s}}\left(\frac{M_{\mathrm{s}}}{M_{\mathrm{r}}}\right)^{2}.
\end{equation}
In terms of the effective learning rate $\alpha$ of \eqref{eq:q-prior}, the law is $\alpha \propto M^{-1}$: the effective learning rate of the head is inversely proportional to the number of steps. For the reported runs, \eqref{eq:q-scaling} gives $q_{\mathrm{r}} = 5.98\times10^{-5}\times(781/5004)^2 = 1.46\times10^{-6}$ (Table~\ref{tab:ft-configs}). The exponent is a consequence of the steady state of the filter and is not fitted to data.

\paragraph{Empirical validation.} We tested \eqref{eq:q-scaling} by measuring the best value of $q$ at three training durations. On DINOv3-Large we evaluated seven values of $q$, six from $1.29\times10^{-7}$ to $1.29\times10^{-2}$, one decade apart, and $4.09\times10^{-3}$, at $M = 781$, $2502$ and $5004$ steps, with one seed per setting. All other hyperparameters were those of the tuned configuration, transferred to each duration by the rule for AdamW above, so that only the number of steps and $q$ varied. The best value on this grid decreases by one decade from $781$ to $2502$ steps and by one further decade from $2502$ to $5004$ steps:

\begin{center}
\begin{tabular}{lccc}
\toprule
steps $M$ & $781$ & $2502$ & $5004$ \\
\midrule
best $q$ & $1.29\times10^{-4}$ & $1.29\times10^{-5}$ & $1.29\times10^{-6}$ \\
\bottomrule
\end{tabular}
\end{center}

Fitting $q \propto M^{-p}$ to the first interval gives $p = 1.98$. The second interval is consistent with $p = 2$ within the resolution of the grid: $p = 2$ predicts $3.2\times10^{-6}$ at $M = 5004$, between the grid values $1.29\times10^{-6}$, which has the lowest test NLL ($0.4212$), and $1.29\times10^{-5}$ ($0.4256$). An exponent $p = 1$ fits less well: it predicts $4.03\times10^{-5}$ at $M = 2502$, between the grid values $1.29\times10^{-5}$ and $1.29\times10^{-4}$, and at the latter the test NLL is $0.011$ higher than at the optimum ($0.4374$ against $0.4267$).

\paragraph{Effect of a value that is too large.} When $q$ is roughly two decades above the value appropriate for the training duration, the in-distribution NLL reaches its minimum near half the budget and increases for the remainder. Below the appropriate value the curve
descends monotonically to its final evaluation instead. This is why $q$ cannot be transferred unchanged from the short duration of the search, and why the search scores the end of the trajectory (Appendix~\ref{sec:ft-search}).

\paragraph{Test on a second backbone.} We applied \eqref{eq:q-scaling} to a
backbone on which it was not measured: DINOv3-Huge, at a matched budget, with the
same tuned trial and only $q$ rescaled. The rescaling reduces the rise of the validation NLL past its minimum by
$65\%$ and improves the test NLL at the validation-optimal evaluation from $0.3935$ to $0.3761$, and at the end of training from $0.4045$ to $0.3760$ (one seed).

\paragraph{Limits.} The derivation of \eqref{eq:q-scaling} uses the scalar reduction of Appendix~\ref{app:filter} in the regime $\lambda q \ll 1$ and assumes a fixed batch size. The exponent is measured at one backbone size. The rise of the validation NLL that remains at DINOv3-Huge is consistent either with a starting value of $q$ that was selected under
an objective that scores the minimum of a truncated curve, or with a dependence of the exponent on the width of the backbone
that we have not identified, and the two are not separated by
the experiments reported here.

\subsection{Final runs}
\label{sec:ft-final}

Table~\ref{tab:ft-configs} gives the values selected by the search and the values after the transfer. Each method is run with the transferred values for five seeds, 15 runs in all. Validation and test metrics are computed every $32{,}768$ samples, which gives $78$ evaluations per run, and the shift sets are scored at every second evaluation.
Cross-entropy and IBProbit evaluate the shift sets in the training process;
VBLL writes checkpoints that a separate worker scores, which is why its compute
is accounted separately in Appendix~\ref{sec:appendix-setup}. All 15 runs completed with every scheduled
evaluation present.

\paragraph{Wall-clock time and memory.} The wall-clock comparison of Figure~\ref{fig:cost-to-threshold} is made on fixed hardware with an identical training protocol across methods. It remains dependent on the implementation (matrix-multiplication dispatch, mixed precision, Cholesky kernel, just-in-time compilation) and should be interpreted as a comparison between these implementations rather than between the methods in general. We do not report peak memory for the fine-tuning runs. At $d = 1281$ and $K = 1000$ the states of the two Bayesian heads (Appendix~\ref{app:cost}) differ by less than $2$\,MB, which a measurement of the peak memory of end-to-end training cannot resolve, because the activations of the backbone dominate the peak. A comparison across frameworks would in addition have to match XLA's \texttt{peak\_bytes\_in\_use} against PyTorch's \texttt{max\_memory\_reserved}, statistics that are defined differently enough that the choice between \texttt{max\_memory\_reserved} and \texttt{max\_memory\_allocated} alone changes the PyTorch result by several GiB, which is more than the differences being compared.

\begin{table}[ht!]
\centering
\caption{Selected fine-tuning configurations at DINOv3-Huge. The upper rows are
the values the search selected at $781$ steps; the lower rows are what each run
of Figures~\ref{fig:sample-efficiency}--\ref{fig:cost-to-threshold} and
Table~\ref{tab:ft-endpoint} actually used at $5{,}004$ steps, after the transfer of Appendix~\ref{sec:appendix-q}.
Every run uses batch size $B = 512$, a constant learning rate and the same
layer-wise decay.}
\label{tab:ft-configs}
\small
\setlength{\tabcolsep}{4pt}
\begin{tabular}{lccc}
\toprule
& AdamW+CE & VBLL & IBProbit \\
\midrule
\multicolumn{4}{l}{\textit{Selected at $781$ steps}} \\
completed trials & $61$ & $59$ & $52$ \\
learning rate / $\sqrt{B}$ & $4.27\times10^{-6}$ & $6.34\times10^{-6}$ & $6.13\times10^{-6}$ \\
weight decay / $\sqrt{B}$ & $1.60\times10^{-7}$ & $2.92\times10^{-4}$ & $1.81\times10^{-5}$ \\
$\beta_1$-horizon / $\beta_2$-horizon & $95.9$ / $0.87$ & $90.7$ / $2.15$ & $84.5$ / $10.7$ \\
head / backbone learning rate & $29.7$ & $479$ & --- \\
KL ratio / prior scale & --- & $5.68\times10^{-4}$ / $533$ & --- \\
process noise $q$ & --- & --- & $5.98\times10^{-5}$ \\
\midrule
\multicolumn{4}{l}{\textit{As run at $5{,}004$ steps}} \\
learning rate & $3.81\times10^{-5}$ & $5.67\times10^{-5}$ & $5.48\times10^{-5}$ \\
head learning rate & $1.13\times10^{-3}$ & $2.72\times10^{-2}$ & --- \\
weight decay & $1.43\times10^{-6}$ & $2.61\times10^{-3}$ & $1.62\times10^{-4}$ \\
$\beta_1$ / $\beta_2$ & $0.9808$ / $0.9998$ & $0.9819$ / $0.9996$ & $0.9831$ / $0.9979$ \\
process noise $q$ / inner iterations $T$ & --- & --- & $1.46\times10^{-6}$ / $256$ \\
\bottomrule
\end{tabular}
\end{table}


\end{document}